\documentclass{egpubl}
\usepackage{pg2025}

\SpecialIssuePaper         

\CGFStandardLicense

\usepackage[T1]{fontenc}
\usepackage{dfadobe}  

\usepackage{cite}  
\BibtexOrBiblatex
\electronicVersion
\PrintedOrElectronic
\ifpdf \usepackage[pdftex]{graphicx} \pdfcompresslevel=9
\else \usepackage[dvips]{graphicx} \fi

\usepackage{egweblnk}

\usepackage{bm}
\usepackage{booktabs}
\usepackage{caption}
\usepackage{multirow}

\usepackage[skins]{tcolorbox}
\usepackage{colortbl}
\usepackage{tabularx}
\usepackage{soul}

\usepackage{subcaption}  
\usepackage{xcolor}      
\usepackage{amsmath}     
\usepackage{amssymb}     
\usepackage{algorithm}   
\usepackage{algpseudocode} 
\usepackage{float}       

\newcommand{\cref}[1]{\ref{#1}}

\newcommand{\ie}{i.e.}

\definecolor{best}{RGB}{253, 104, 100}
\definecolor{second}{RGB}{255, 204, 153}
\newtcbox\best{hbox, on line, colback=best, enhanced, frame hidden, boxrule=0pt, top=-2pt, bottom=-2pt, right=-2pt, left=-2pt, sharp corners}
\newtcbox\second{hbox, on line, colback=second, enhanced, frame hidden, boxrule=0pt, top=-2pt, bottom=-2pt, right=-2pt, left=-2pt, sharp corners}

\definecolor{egblue}{rgb}{0.21,0.49,0.74}
\hypersetup{pagebackref,breaklinks,colorlinks,allcolors=egblue}

\title[Joint Deblurring and 3D Reconstruction]%
      {Joint Deblurring and 3D Reconstruction for Macrophotography}

\author[Yifan Zhao et al.]
{\parbox{\textwidth}{\centering Yifan\,Zhao$^{1}$\orcid{0009-0000-5499-4459} and Liangchen\,Li$^{1}$\orcid{0009-0004-8176-0745} and Yuqi\,Zhou$^{1}$\orcid{0009-0009-5656-8475} and Kai\,Wang$^{2}$\orcid{0000-0002-1171-0281} and Yan\,Liang$^{1}$\orcid{0009-0009-2048-1837} and Juyong\,Zhang$^{1}$\orcid{0000-0002-1805-1426}\thanks{Corresponding Author}
        }
        \\
{\parbox{\textwidth}{\centering $^1$University of Science and Technology of China\\
\vspace{0.5em}
         $^2$China Unicom
       }
}
}

\begin{document}


\twocolumn[{%
\renewcommand\twocolumn[1][]{#1}%
\maketitle
\begin{center}
    \centering
    \captionsetup{type=figure}
    \begin{tabular}[b]{cc}

        \begin{subfigure}[b]{0.42\linewidth}
            \includegraphics[width=\textwidth]{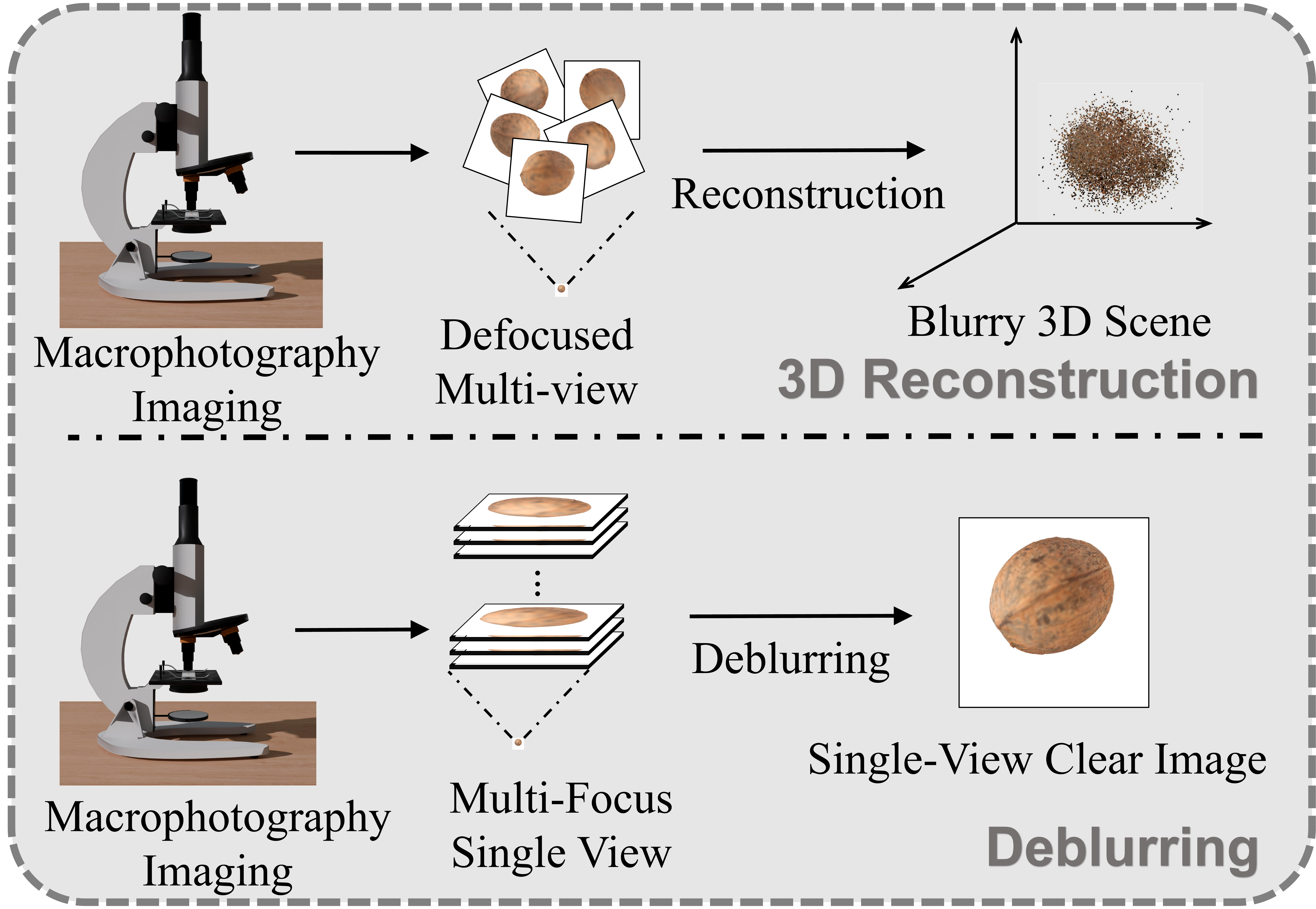}
        \end{subfigure} &
        \begin{subfigure}[b]{0.56\linewidth}
            \includegraphics[width=\textwidth]{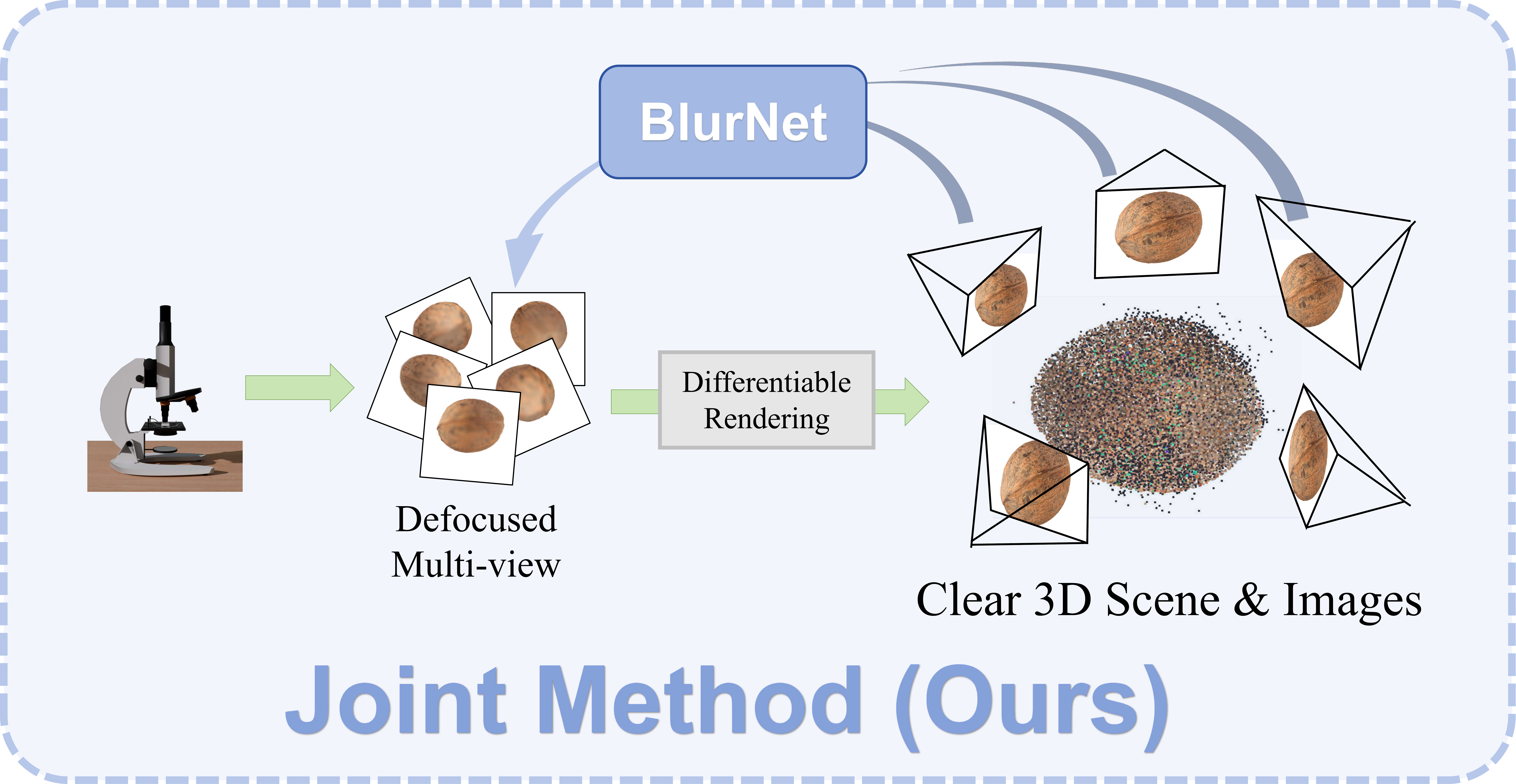}
        \end{subfigure}  \\

        (a) Traditional Methods  & (b) Our Method \\
    \end{tabular}
    \caption{Defocus blur is a long-standing problem in macrophotography. Traditional multi-focus image fusion deblurring methods usually require the acquisition of a large number of input images. Without deblurring, 3D reconstruction of tiny objects can only produce blurred 3D scenes. To address this problem, we propose a method for combining deblurring and 3D reconstruction of macrophotography, which can produce clear 3D scenes and images.}
    \vspace{2em}
    \label{fig:teaser}
\end{center}%
}]


\begin{abstract}
   Macro lens has the advantages of high resolution and large magnification, and 3D modeling of small and detailed objects can provide richer information. However, defocus blur in macrophotography is a long-standing problem that heavily hinders the clear imaging of the captured objects and high-quality 3D reconstruction of them. Traditional image deblurring methods require a large number of images and annotations, and there is currently no multi-view 3D reconstruction method for macrophotography. In this work, we propose a joint deblurring and 3D reconstruction method for macrophotography. Starting from multi-view blurry images captured, we jointly optimize the clear 3D model of the object and the defocus blur kernel of each pixel. The entire framework adopts a differentiable rendering method to self-supervise the optimization of the 3D model and the defocus blur kernel. Extensive experiments show that from a small number of multi-view images, our proposed method can not only achieve high-quality image deblurring but also recover high-fidelity 3D appearance.

\ccsdesc[500]{Computing methodologies~Computer graphics}
\ccsdesc[300]{Computing methodologies~Image processing}
\ccsdesc[300]{Computing methodologies~3D imaging}

\printccsdesc   
\end{abstract}  
\section{Introduction}
\label{sec:introduction} 

Macrophotography~\cite{ma_comprehensive_2021,cuny_live_2022,masters2008history} is a photographic technique that uses a specially designed macro lens or even a low-power microscope to photograph tiny objects, such as insects and mineral particles, at a sub-centimeter size. Unlike microscopy which typically deals with microscopic specimens at micrometer scale, macrophotography focuses on slightly larger subjects (millimeter to centimeter scale) and employs different optical systems. This technique enables researchers and artists to observe and document the subtle world that is difficult to observe with the naked eye but visible in the macro world, creating possibilities for high-fidelity 3D reconstruction of these small objects. Applications of macrophotography span diverse fields such as entomological documentation~\cite{biss2017microsculpture, app12020769}, precision industrial inspection~\cite{RODRIGUEZMARTIN201554, JEFFREYKUO2017248}, and artistic expression~\cite{Orci2002, prakash2024}.




     
    


\begin{figure*}[htbp]
    \centering
    \includegraphics[width=\linewidth]{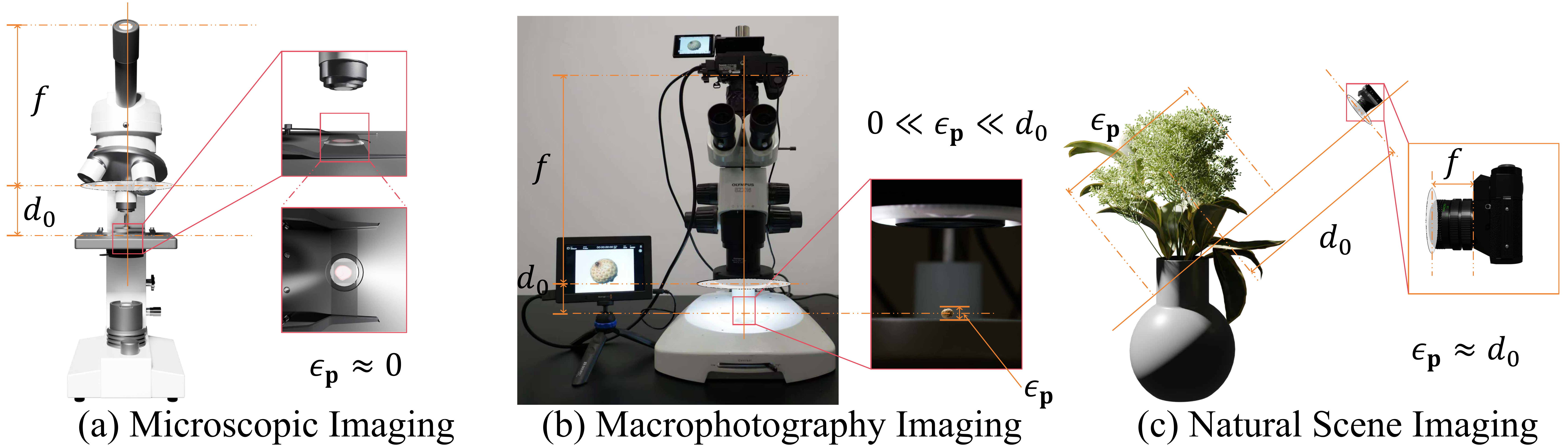}

    \vspace{1em}

    \resizebox{\linewidth}{!}{
        \begin{tabular}{l |ccc}
            Imaging Scene & Microscopic Scenes & Macrophotography Scenes & Natural Scene \\
            \midrule[1pt]
            Object Size & Micrometer scale & Sub-centimeter scale & Centimeter to meter scale \\
            Complexity of 3D Information & 2D plane & Rich 3D details & Rich 3D details  \\
            DoF(Depth of Field) & Shallow DoF & Shallow DoF & Wide DoF  \\
        \end{tabular}
    }

    \caption{
        \textbf{Comparison of imaging scenes and their characteristics.} In microscopic imaging, imaged objects are often slides with no 3D reconstruction possibilities. In macrophotography imaging, the imaged objects are with 3D structure, but small and have a shallow depth of field, resulting in defocus blur. In natural imaging, objects are distributed over a wide range of depths. During real macrophotography image capture, one of our measurements is \(f+d_0\approx480\text{mm}\) and \(\epsilon_{\mathbf{p}}\approx3\text{mm}\). This critical difference makes macro scenes 3D reconstructable but extremely sensitive to depth variations compared to other scenes, which the proposed method aims to address.
        }
    \label{fig:natural&micro}
\vspace{-1em}
\end{figure*}

As illustrated in Fig.~\ref{fig:natural&micro}, macrophotography operates in a unique optical regime that distinguishes it from other imaging process. In terms of optical characteristics, macro lenses typically have large focal lengths and operate at close working distances, creating fundamental differences in image formation. Compared to normal photography, macrophotography works with a much smaller subject-to-camera distance and have a much smaller depth of field, magnifying subjects to near life-size or greater on the image sensor. In contrast to microscopy, which utilizes compound optical systems with transmitted light and fixed working distances, macrophotography employs specialized lenses with reflected light and variable working distances. While microscopes achieve very high magnifications (typically 40-1000×), macrophotography operates in a lower magnification range (1-5×) but offers greater flexibility in composition, lighting control, and subject placement. These distinct optical properties create a specific imaging environment with its own set of technical challenges.

The principal challenge in macrophotography arises from its extremely shallow depth of field. Due to the close focusing distances and relatively large apertures needed for adequate light gathering, even tiny variations in subject depth result in significant focus disparities. This creates a severe technical limitation: it becomes nearly impossible to capture an entire three-dimensional object in sharp focus with a single exposure. This limitation, known as defocus blur, manifests much more prominently in macrophotography than in normal photography and follows different optical patterns than in microscopy. When attempting to create 3D reconstructions from macrophotography images, this defocus blur causes significant image distortion and detail loss, degrading quality and hindering high-fidelity results. Consequently, specialized deblurring techniques designed specifically for macrophotography's unique optical characteristics become essential.

As shown in Fig.~\ref{fig:teaser}, a large number of works have been proposed to address the defocus blur problem~\cite{Zhang_2024_CVPR_mptcatablur, jiang_blind_2020,ghamsarian_deblurring_2020,zhao_new_2020,wang_deblurring_2023,zhang_high-throughput_2023,kornilova_deep_2021,liu_deep_2020}, but they usually require large number of captured images and are only applicable to deblurring in 2D image space. They cannot aggregate information from adjacent views and ensure multi-view consistency. On the other hand, existing 3D scene deblurring methods do not consider the difference between macrophotography images and natural images. To the best of our knowledge, no one has considered the problem of restoring the clear 3D appearance of detailed tiny objects from blurred inputs. Compared with natural scenes, macro scenes are more sensitive to parameters, have shallower depth of field, and larger defocus blur, making reconstructing macro scenes more challenging.

In this work, we propose the first joint self-supervised optimization method that can simultaneously obtain high-quality 3D scene and clear images for subcentimeter tiny objects. We observe that the essence of imaging is the projection of a clear 3D scene onto the imaging plane, and the depth variation during the projection process produces defocus blur. Therefore, our key idea is to accurately simulate the imaging process of blur caused by depth variation in the macro scene in the reconstructed 3D scene space. As long as we get a blurred image consistent with the input along this process, we can complete a high-fidelity 3D reconstruction of the macro scene through the latest differentiable rendering methods\cite{kerbl20233dgs}. This 3D to 2D deblurring method achieves clear 3D scene for multi-view rendering while maintaining the geometric features of the image.

To simulate the blurred imaging process, we study the characteristics of macrophotography. Our key observation shows that the defocus blur in macrophotography images is much more sensitive to depth changes than natural images. To this end, we carefully design a blur module. This module contains multiple networks to fully extract information such as depth in 3D scenes and generate defocus maps. These processes take into account the characteristics of macrophotography and effectively avoid unacceptable results caused by small errors in parameter estimation. Based on the defocus map, we assign a blurred convolution kernel to each pixel to restore a clear 2D macrophotography image. In addition, we introduce the latest differentiable rendering method~\cite{kerbl20233dgs} into our method as a reliable 3D representation to estimate depth, maintain multi-view consistency and restore the 3D appearance of small objects. Combining it with the blur module achieves the simulation of the blurred imaging process, so that we only use the blurred image as input and obtain clear 3D reconstruction results through self-supervised optimization.

We conduct extensive experiments on both synthetic and real datasets of macrophotography defocus blur. Results show that our method outperforms existing single image deblurring methods (\ie, combined with 3D reconstruction methods) and other existing 3D scene deblurring methods in terms of deblurred reconstruction, as shown in Sec.~\ref{sec:experiments}. In summary, our contributions can be summarized into three parts:

\begin{itemize}
\item[$\bullet$] We propose the first joint self-supervised deblurring and 3D reconstruction method for macrophotography, recovering the sharp 3D appearance of subcentimeter-sized small objects from blurry inputs without requiring additional supervision.
\item[$\bullet$] We formulate the deblurring problem within a multi-view 3D reconstruction framework, leveraging 3D constraints across different viewpoints to overcome the ill-posed problem of single-image deblurring and enable joint optimization for both clear 3D scenes and accurate blur modeling.
\item[$\bullet$] We design BlurNet to extract depth and other features from 3D scenes to guide reconstruction, effectively incorporating optical priors of macrophotography to enhance deblurring performance in macro scenes.
\end{itemize}


\section{Related Work}
\label{sec: related work}

\subsection{Focusing Techniques for Macrophotography Devices}

In order to handle the defocus blur in macrophotography images, various specialized capturing devices have been designed to mitigate this effect~\cite{xu2017wavefront,pinkard2019deep,li2021deep, trukhova2022microlens,yoo20183d,lin2019multi}. Auto-focusing microscopes~\cite{xu2017wavefront,pinkard2019deep,li2021deep} attempt to detect the depth of objects in the field of view and drive the focusing device to adjust the optical system to complete the focusing process~\cite{zhang2021}. Microlens-assisted microscopes~\cite{trukhova2022microlens} replace a single lens in a microscope with a microlens array, thereby expanding the field of view. Multi-focus microscopes~\cite{lin2019multi} can set multiple focal points at the same time, but at the cost of a complex system and reduced resolution and speed. However, defocus blur is caused by the inherent optical effects of the lens and is difficult to eliminate, so it is necessary to introduce an image defocus deblurring method in the post-processing of macrophotography.

\subsection{Defocus Deblurring for Macrophotography}

In macrophotography images, defocus blur is usually directly modeled using a complex network to map the blurred image to the sharp image. Due to the high sensitivity of these images to parameters, end-to-end methods have higher robustness and higher performance. Based on this, many works~\cite{Yang2018,yun2023, Zhang_2024_CVPR_mptcatablur,wang2023} have studied the removal of defocus blur in macrophotography images. For example, early methods~\cite{Yang2018} attempted to use deep neural networks to evaluate the focus quality of macrophotography images. Later, a coarse-to-fine deblurring method~\cite{wang2023} based on U-Net~\cite{olaf2015unet} was shown to be effective. The single focus estimation method KDAF~\cite{yun2023} introduced the concept of blur kernel to predict the focal length of macrophotography images. Recently, MPT-Catablur~\cite{Zhang_2024_CVPR_mptcatablur} introduced an attention mechanism and used a pyramid structure to achieve image deblurring. These methods usually require training a powerful and large restoration model, which contains a huge number of training pairs and a large neural network, consuming a lot of training time and memory. However, due to the difficulty in obtaining real training pairs on a large scale, such a model is difficult to adapt to every type of degradation. The 2D image spatial deblurring method is not easy to apply to 3D scenes due to the inevitable lack of multi-view consistency.

\subsection{3D Reconstruction from Defocus Blur}

Differentiable rendering techniques offer more possibilities for defocus deblurring. 3D Gaussian Splatting (3DGS)~\cite{kerbl20233dgs} and Neural Radiance Field (NeRF)~\cite{mildenhall2020nerf} are effective methods for creating high-fidelity 3D scenes from 2D images. These techniques introduce multi-view consistent 3D spatial information into the image, enabling deblurring methods to take it into account. Many deblurring methods~\cite{peng2024bags,lee2024deblurring3dgs,wang2024dofgsadjustabledepthoffield3d,peng2023pdrf,Ma_2022_CVPR_deblurnerf} based on NeRF~\cite{mildenhall2020nerf} and 3DGS~\cite{peng2024bags,lee2024deblurring3dgs,wang2024dofgsadjustabledepthoffield3d,peng2023pdrf,Ma_2022_CVPR_deblurnerf} attempt to reconstruct sharp 3D scenes from multi-view images with defocus blur. For example, Deblur-NeRF~\cite{Ma_2022_CVPR_deblurnerf} and PDRF~\cite{peng2023pdrf} use 2D sparse pixel-level kernels to predict blurred images. With the development of 3DGS, 3DGS-based deblurring methods have also emerged. Deblurring-3DGS~\cite{lee2024deblurring3dgs} optimizes the spatial properties of Gaussian primitives to obtain latent sharp 3D scenes, BAGS~\cite{peng2024bags} uses the efficient rasterizer of 3DGS to estimate the blur kernel pixel by pixel, and DOF-GS~\cite{wang2024dofgsadjustabledepthoffield3d} uses the physical model of the camera for controllable depth of field rendering. Although these 3D deblurring methods successfully recover sharp images from blurry inputs, there is currently no work that attempts to extend the applicability of deblurring methods to macrophotography images. Therefore, we propose a joint deblurring and 3D reconstruction method for these tiny objects, which is able to recover sharp 3D appearance and images from a small number of multi-view blurred macrophotography images, considering the differences between macro and natural scenes.

\section{Method}
\label{sec:method}

Our goal is to recover the geometry and appearance of a macro scene \(\mathcal{S}\) from multi-view blurred images \( \{ \mathbf{I}_m \}_{m=1}^M\) to produce sharp images. The pipeline of our method is shown in Fig.~\ref{fig:pipeline}. To achieve this goal, we design an end-to-end optimization method, which explicitly models the defocus blur kernel in the micro-scene and utilizes a small network to predict it, jointly optimize the network and the 3D scene, and finally obtain a sharp 3D scene without complex network architecture and supervision.

In Sec.~\ref{subsec:theory}, we analyze the modeling of the defocus blur kernel in macrophotography and how to build a network to predict it. Based on this, we propose a joint approach for defocus deblurring and 3D reconstruction in Sec.~\ref{subsec:framework}. Sec.~\ref{subsec:strategy} details the optimization techniques of our approach.

\subsection{Defocus Blur Kernel in Macrophotography}
\label{subsec:theory}

In many scenarios, the impact of defocus blur on imaging results cannot be ignored, which can be measured by the Circle of Confusion (CoC). Assuming that the camera uses an ideal thin lens, the CoC of a pixel \(\mathbf{p}\) with a depth of \(d_\mathbf{p}\) refers to the circular area in the imaging plane onto which the pixel is projected by the lens. According to the imaging law in geometric optics~\cite{li2005applied, hecht2017optics}, the radius \(\sigma_\mathbf{p}\) of pixel \(\mathbf{p}\) satisfies
\begin{equation}
  \sigma_\mathbf{p}=\frac{1}{2}fA\frac{|d_\mathbf{p}-d_0|}{d_\mathbf{p}(d_0-f)},
  \label{eq4}
\end{equation}
where \(f\) is the focal length of the camera, \(A\) is the F-number of the aperture, and \(d_0\) is the depth of focal plane. 

According to Eq.(\ref{eq4}), large focal lengths and apertures of macro lenses and small depth of the object drastically increase blur-depth sensitivity. The equivalent parameters of cameras with macro lenses exceed normal cameras by orders of magnitude. In addition, the depth of the object in macro scenes is much smaller than that in natural scenes. These factors yield a critically shallow depth of field. It fundamentally prevents full focus coverage and causes unavoidable defocus blur regardless of focal plane adjustments. More details about CoC in macrophotography can be found in the Supplementary Material.

Given that the defocus blur in macrophotography images exhibits strong isotropy and smoothness~\cite{quan2021}, we consider the CoC of the pixel \(\mathbf{p}\) as an isotropic 2D Gaussian blur kernel~\cite{Shi2015JustND,xu2017estimating, Lee2019DeepDM} \(\mathbf{G}_{\mathbf{p}}\) with variance \(\sigma_{\mathbf{p}}\) in practice. Therefore, consider a defocused macrophotography image \(\widetilde{\mathbf{I}}\), and its corresponding clear image \(\mathbf{I}\), the color of pixel \(\mathbf{p}\) in two images is related by the following formula:

\begin{figure*}[t]
\centering
\includegraphics[width=0.96\linewidth]{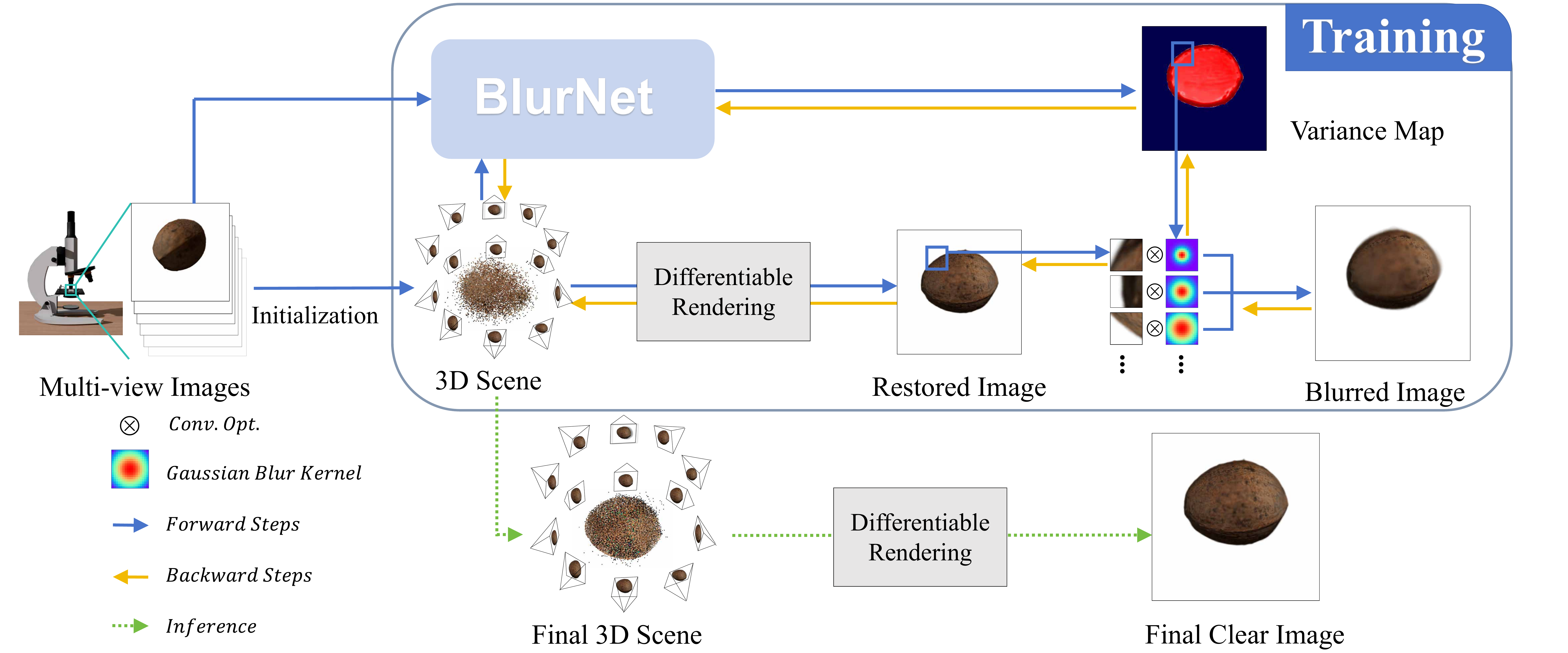}
\vspace{-0.2cm}
\caption{\textbf{An overview of our joint deblurring and 3D reconstruction method.} First, we initialize a differentiable coarse 3D scene from multi-view input blurred images. Afterwards, BlurNet leverages information from the input images to generate a variance map (\ie defocus map) for each view. The variance map is then convolved with the rendered image pixel-wise. The output is a correctly blurred image, supervised by the blurred input images during training. Note that this process ultimately produces a sharp scene.}
\vspace{-1.5em}
\label{fig:pipeline}
\end{figure*}

\begin{equation}
  \widetilde{\mathbf{I}}\left[\mathbf{p}\right] = \sum_i\sum_j \mathbf{G}_{\mathbf{p}}\left[i,j\right]\mathbf{I}\left[\mathbf{p}+(i,j)\right],
  \label{eq2}
\end{equation}
where \(\mathbf{G}_{\mathbf{p}}\) is the blur kernel of pixel \(\mathbf{p}\), satisfying
\begin{equation}
  \mathbf{G}_{\mathbf{p}}\left[i, j\right]= \frac{\exp(-(i^2+j^2)/2\sigma_\mathbf{p}^2)}{\sum_{i,j=-K}^{K}\exp(-(i^2+j^2)/2\sigma_\mathbf{p}^2)}.
  \label{eq3}
\end{equation}

As shown in Fig.~\ref{fig:natural&micro}, our key observation is that in macro scenes, while the possibility of 3D reconstruction is preserved compared to microscopic scenes, imaged objects are often closer to the camera focal plane than in natural scenes, and the depth \(d_{\mathbf{p}}\) of each pixel \(\mathbf{p}\) varies only within a very small range around the focal plane depth \(d_0\). Even small depth perturbations can lead to visible defocus blur. Due to the sensitivity of \(\sigma_\mathbf{p}\) to small errors, we choose to extract depth and other features through several carefully designed networks and predict \(\sigma_\mathbf{p}\) from them.

Denote \(\epsilon_{\mathbf{p}}\) as the depth of pixel \(\mathbf{p}\) relative to the focal plane, i.e. \(d_{\mathbf{p}} - d_0 = \epsilon_{\mathbf{p}}\). Then Eq.~\eqref{eq4} can be written as
\begin{equation}
    \sigma_\mathbf{p}=\frac{1}{2}fA\frac{|\epsilon_{\mathbf{p}}|}{(d_0+\epsilon_{\mathbf{p}})(d_0-f)}.
    \label{eq5}
\end{equation}

Given that in macrophotography the depth of the object \(d_\mathbf{p}\) is typically very close to the focal plane depth \(d_0\), the relative depth \(|\epsilon_\mathbf{p}| = |d_\mathbf{p} - d_0|\) is much smaller than \(d_0\) itself (i.e., \(|\epsilon_\mathbf{p}| \ll d_0\)).
This crucial condition allows for the approximation \( (d_0 + \epsilon_\mathbf{p}) \approx d_0 \) in the denominator of Eq.~\eqref{eq5}.
Consequently, Eq.~\eqref{eq5} simplifies to:
\begin{equation}
  \sigma_\mathbf{p} \approx \frac{fA}{2} \frac{|\epsilon_\mathbf{p}|}{d_0(d_0-f)}.
  \label{eq_intermediate_approximation} 
\end{equation}
By defining a term \(\alpha = \frac{fA}{2d_0(d_0-f)}\), which groups the camera's optical parameters (\(f, A\)) and the focal plane depth (\(d_0\)) for the current view, we arrive at the factored form:
\begin{equation}
  \sigma_\mathbf{p}=\alpha\cdot|\epsilon_\mathbf{p}|.
  \label{eq6}
\end{equation}

Eq.~\eqref{eq6} shows that \(\sigma_\mathbf{p}\) can be divided into two parts \(\alpha\) and \(\epsilon_{\mathbf{p}}\), where \(\epsilon_{\mathbf{p}}\) is relative to \(d_{\mathbf{p}}\) but \(\alpha\) is not. Therefore, we design two networks for each part. Formally:
\begin{equation}
  \boldsymbol{\sigma}_{\mathbf{p}} = \boldsymbol{A}\odot \boldsymbol{E}_{\mathbf{p}},
  \label{eq7}
\end{equation}
where \(\boldsymbol{E}\) is a feature map relative to depth and \(\boldsymbol{A}\) is the feature independent to depth, both of which are output by a carefully designed network, and \(\odot\) represents element-wise multiplication. The network architecture is detailed in Sec.~\ref{subsec:framework}.

\subsection{Joint Method for Deblurring and Reconstruction}
\label{subsec:framework}

As shown in Fig.~\ref{fig:pipeline}, the input of our method is a set of multi-view blurry macrophotography images \(\{ \mathbf{I}_m \}_{m=1}^M\). Through the latest 3D reconstruction algorithm~\cite{kerbl20233dgs, misc_colmap1, misc_colmap2}, we get a roughly initialized 3D scene, which can achieve differentiable rendering of novel view images. Then, our method generates correct blurred images by assigning blur kernels to the rendered images through our carefully designed blur kernel network (called BlurNet). Through BlurNet, we can obtain blurred images \(\widetilde{\mathbf{I}}_{\text{out}}\) and clear images \(\mathbf{I}_{\text{out}}\).

Our optimization strategy is to refine the parameters of both the 3D scene \(\mathcal{S}\) which implicitly determines \(\mathbf{I}_{\text{out}}\) and the depth map \(\mathbf{D}\) rendered from the scene, and BlurNet. This joint optimization is supervised by minimizing the discrepancy between the input blurry image \(\mathbf{I}\) and the simulated blurred image \(\widetilde{\mathbf{I}}_{\text{out}}\). The ultimate goal is to recover a high-fidelity clear 3D scene \(\mathcal{S}\) and a BlurNet capable of accurately modeling the specific defocus characteristics in macrophotography. This process ensures that the information about the defocus blur is effectively captured by BlurNet, while the 3D scene representation converges towards a sharp depiction of the object.

\begin{figure}
    \centering
    \includegraphics[width=\linewidth]{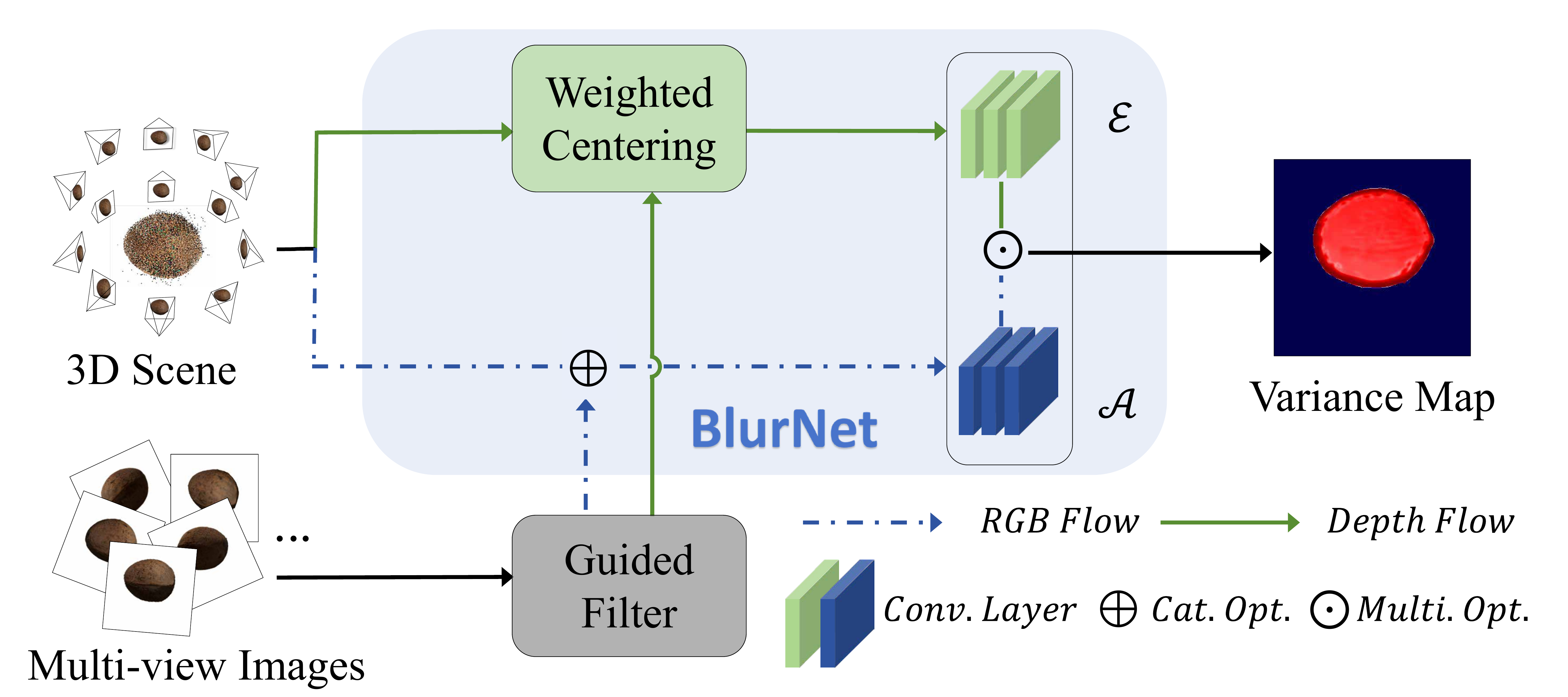}
    \vspace{-20pt}
    \caption{\textbf{The architecture of BlurNet.} At the beginning of training, the clarity mask is generated from the input multi-view images via guided filters. The depth map is weighted centered using the clarity mask to generate a relative depth map for the depth CNN \(\mathcal{E}\). The rendered image is concatenated with the clarity mask as input to the RGB CNN \(\mathcal{A}\). The outputs of both CNNs are multiplied to generate a variance map.}
    \label{fig:blurnet}
    \vspace{-1.5em}
\end{figure}

The key component in our method is BlurNet, whose structure is shown in Fig.~\ref{fig:blurnet}. As described in Eq.~\eqref{eq7}, we design two networks to predict \(\sigma_\mathbf{p}\), one of which is related to the depth of the pixel and the other is independent of it. Using the depth map \(\mathbf{D}\) and the rendered RGB image \(\mathbf{I}_{\text{out}}\) output by the 3D scene representation as input of the network, Eq.~\eqref{eq7} can be written as:
\begin{equation}
  \boldsymbol{\sigma}_{\mathbf{p}} = \boldsymbol{A}(\mathbf{I}_{\text{out}})\odot \boldsymbol{E}_{\mathbf{p}}(\mathbf{D}).
  \label{eq8}
\end{equation}

For \(\boldsymbol{E}(\cdot)\) in Eq.~\eqref{eq8}, the original depth map \(\mathbf{D}\) is not enough for extracting depth features. We use the clarity mask \(\mathbf{M}\) as a weight to weighted center the depth map \(\mathbf{D}\). In this way, we can accurately locate the in-focus area in each input image \(\mathbf{I}\), and then calculate the average depth of the in-focus area as \(d_0\) in Eq.~\eqref{eq4}, and finally obtain a relative depth map \(\hat{\mathbf{D}}\). It can be written as:
\begin{equation}
  \hat{\mathbf{D}}_\mathbf{p} = \bigg|\mathbf{D}_\mathbf{p}-\sum_{\mathbf{x}\in \mathcal P}\mathbf{D}_\mathbf{x}\mathbf{M}_\mathbf{x}\bigg|,
  \label{eq9}
\end{equation}
where \(\mathcal{P}\) is the set that contains all pixels in the image \(\mathbf{I}\).

While the global depth map from 3DGS may exhibit inconsistencies, it preserves local geometric fidelity, with neighboring pixels maintaining relative depth relationships critical for blur estimation. Therefore, we design a lightweight CNN encoder \(\mathcal{E}(\cdot)\) operating in these local neighborhoods, learning to extract and refine depth-aware features based on local geometric fidelity, and utilizing the positional encoding technique~\cite{mildenhall2020nerf, matthew2020} to provide a higher dimensional input to the CNN. The structure of the CNN is:
\begin{equation}
  \boldsymbol{E}(\mathbf{D}) = \mathcal{E}(P_L(\hat{\mathbf{D}})),
  \label{eq10}
\end{equation}
where \(P_L(\cdot)\) is a positional encoding function with frequency \(L\) and \(\mathcal{E}\) is a small CNN.

For \(\boldsymbol{A}(\cdot)\) in Eq.~\eqref{eq8}, we also apply a small CNN to extract RGB features, and we find \(\mathbf{M}\) is helpful in feature extraction. Therefore,
\begin{equation}
  \boldsymbol{A}(\mathbf{I}_{\text{out}}) = \mathcal{A}(\mathbf{I}_{\text{out}}\oplus\mathbf{M}),
  \label{eq11}
\end{equation}
where \(\mathcal{A}\) is another small CNN.

BlurNet effectively extracts depth and other necessary features from the 3D scene and original multi-view images, ensuring the robustness of our method. Through BlurNet, we can get a variance map \(\boldsymbol{\sigma}\) as the input of Eq.~\eqref{eq3}, and then we can apply Eq.~\eqref{eq3} and Eq.~\eqref{eq2} to get the blur kernel of all pixels and blurred image \(\widetilde{\mathbf{I}}_{\text{out}}\).

\subsection{Representation and Optimization}
\label{subsec:strategy}

\subsubsection{3D scene representation.}

Our method uses 3DGS~\cite{kerbl20233dgs} as the basic representation of macrophotography scenes. In 3DGS, the scene is represented by a set of Gaussian points \(\{\mathcal{G}_n|n=1,\cdots, N\}\), each of which contains multiple attributes, including center \(x_n\), covariance matrix \(\Sigma_n\), opacity \(\alpha_n\), and color \(\mathbf{c}_n\). With carefully designed tiling, 3DGS provides sufficient rendering speed to develop new methods for clearer multi-view high-fidelity rendering. 
However, practical applications of 3DGS in macrophotography can be challenging. Despite its impressive performance on curated datasets, 3DGS requires high quality images input to work well. Without this, 3DGS often generates undesired Gaussians to overfit observation noise, leading to worse renderings~\cite{peng2024bags}.

\subsubsection{Clarity mask.}
The clarity mask \(\mathbf{M}\) accurately locates the focus area in our method. In the preprocessing, an effective guided filter-based algorithm~\cite{QIU2019} is introduced to obtain the clarity mask \(\mathbf{M}\) for each image input \(\mathbf{I}\).

\begin{figure}[htb!]
    \setlength{\abovecaptionskip}{3pt}
    \setlength{\tabcolsep}{1pt}
    \centering
    \small
    \begin{tabular}[b]{cccc}
        
        \begin{subfigure}[b]{0.25\linewidth}
            \includegraphics[width=\textwidth]{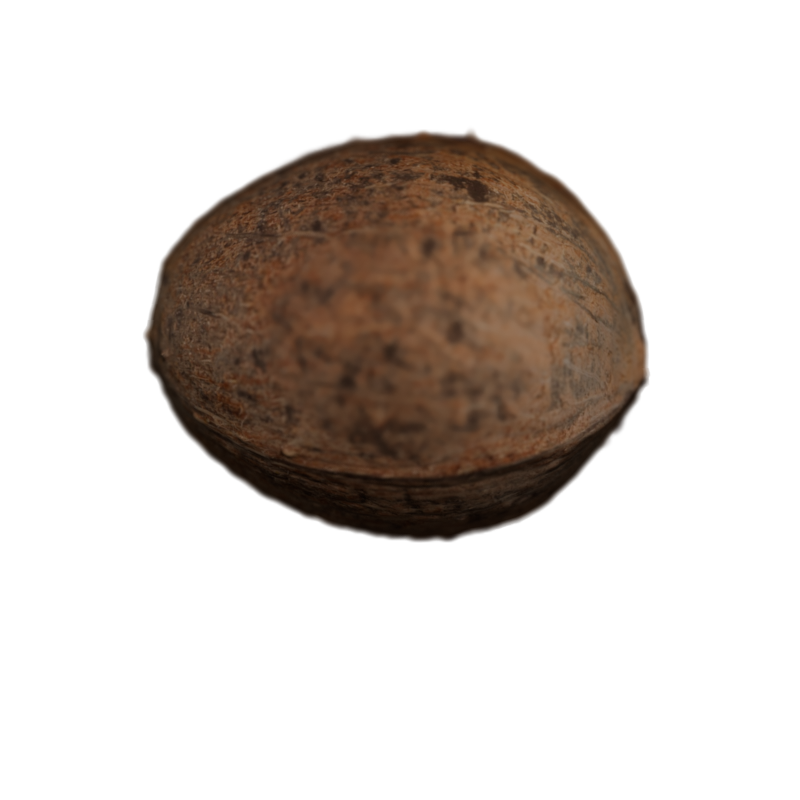}
        \end{subfigure} &
        \begin{subfigure}[b]{0.25\linewidth}
            \includegraphics[width=\textwidth]{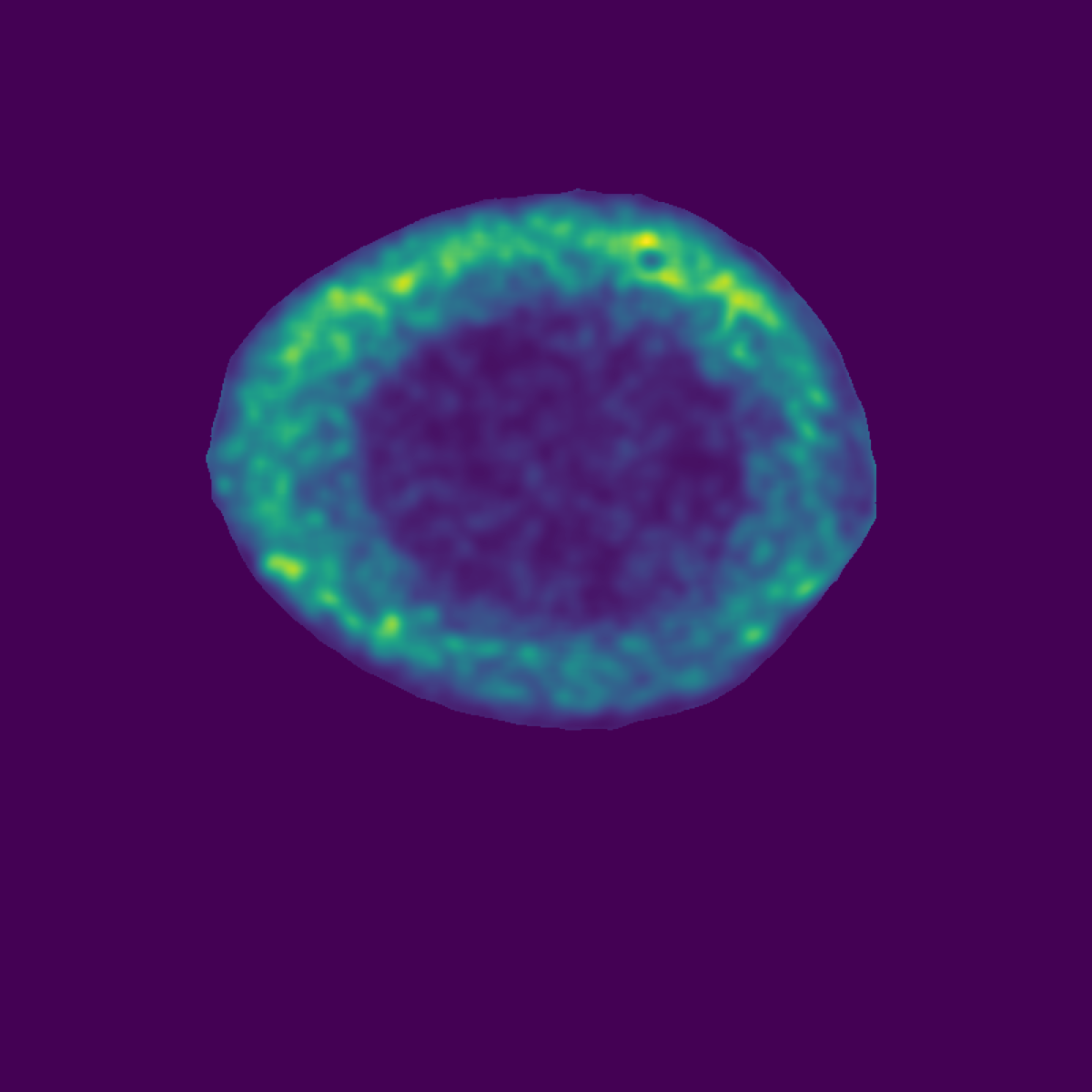}
        \end{subfigure} &
        \begin{subfigure}[b]{0.25\linewidth}
            \includegraphics[width=\textwidth]{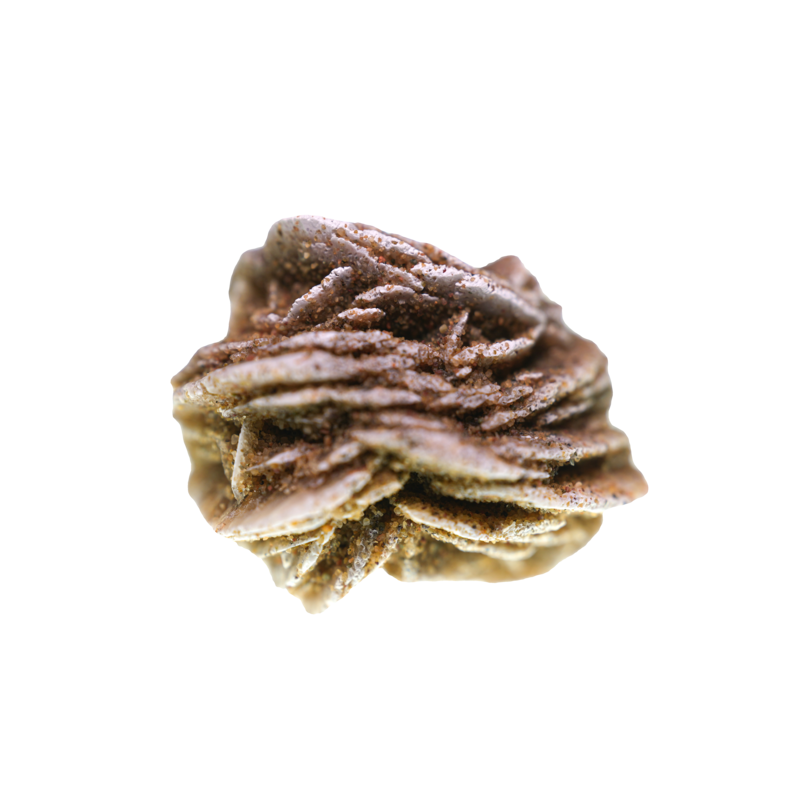}
        \end{subfigure} &
        \begin{subfigure}[b]{0.25\linewidth}
            \includegraphics[width=\textwidth]{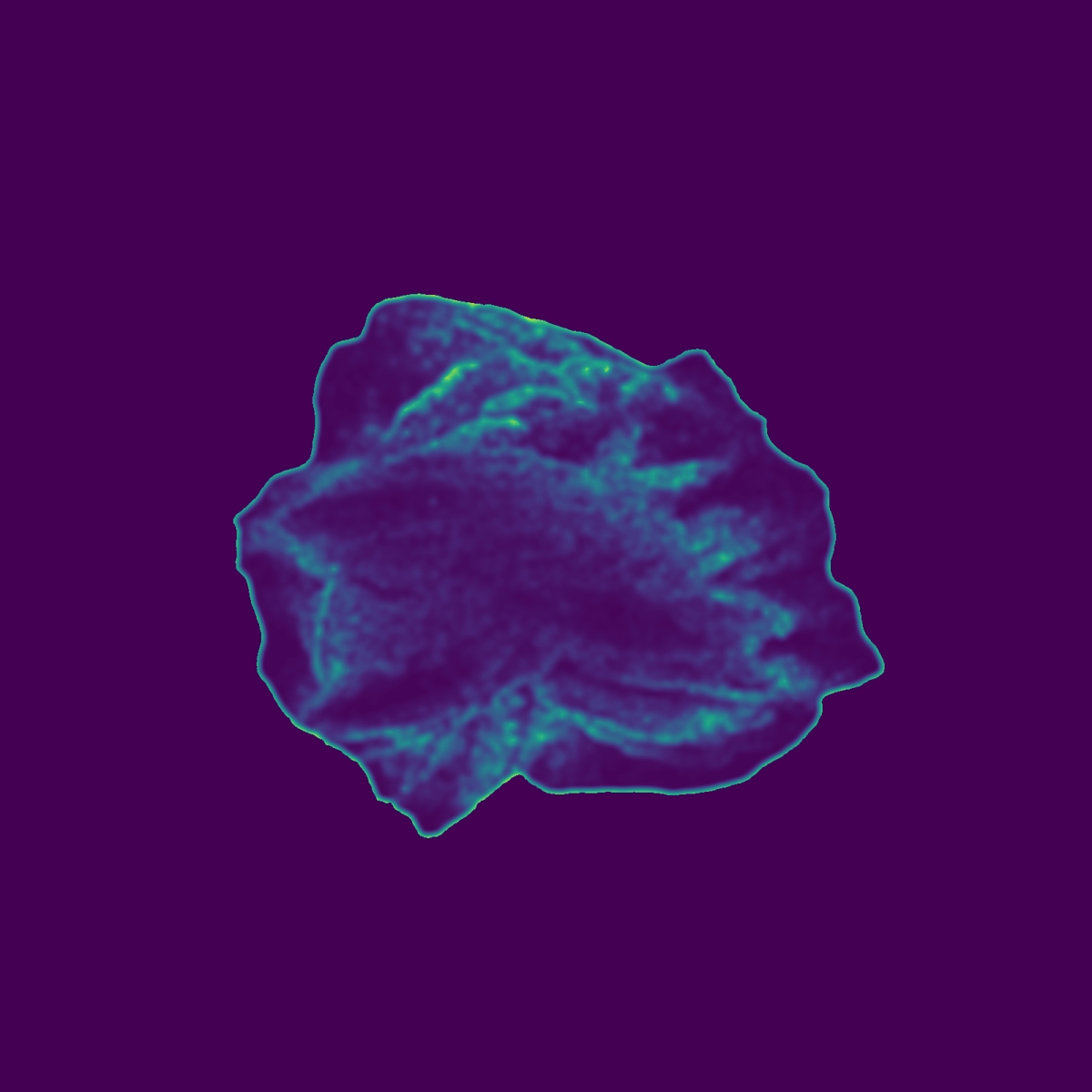}
        \end{subfigure} \\
        Defocus & Clarity Mask & Defocus & Clarity Mask \\
        
        \multicolumn{2}{c}{Synthetic Data} & \multicolumn{2}{c}{Real Data} \\
        
    \end{tabular}
    \vspace{-1em}
    \caption{\textbf{Visualization of computed clarity masks}. Comparison between defocused input and their corresponding clarity masks demonstrates how our method identifies high-frequency, in-focus regions (bright areas in mask) to estimate the focal plane depth.}
    \label{fig:clarity_mask_visualization}
    \vspace{-1em}
\end{figure}

Specifically, at the beginning of training, we use a mean filter to extract the high-frequency information retained in the image to form a coarse clarity mask. Then, with the original image as a guide, we apply a guided filter to the coarse clarity mask to enhance the high-frequency information in the coarse clarity mask to form a refined clarity mask
\begin{equation}
  \mathbf{M}=\text{GF}(\mathbf{I},|\mathbf{I}-\text{MF}(\mathbf{I})|),
  \label{eq12}
\end{equation}
where \(\text{MF}\) is the mean filter operator and \(\text{GF}\) is the guided filter operator. This operation only needs to be performed once at the beginning of training, so it does not slow down training. Fig.~\ref{fig:clarity_mask_visualization} illustrates how clarity masks accurately identifies high-frequency, in-focus regions across blurry input. For more details of the guided filter-based algorithm, please refer to the Supplementary Materials.

\begin{table*}[htbp]
    \centering
    \resizebox{0.98\linewidth}{!}{

    \begin{tabular}{l |cccccccccccccccccc}

     Methods  & \multicolumn{3}{c}{3DGS \cite{kerbl20233dgs}}  & \multicolumn{3}{c}{Deblur-3DGS \cite{lee2024deblurring3dgs}} &  \multicolumn{3}{c}{BAGS \cite{peng2024bags}}  & \multicolumn{3}{c}{Restormer+3DGS \cite{Zamir2021Restormer}} & \multicolumn{3}{c}{INIKNet+3DGS \cite{quan2023}}  & \multicolumn{3}{c}{Ours}  \\

     Metrics & PSNR & SSIM & LPIPS & PSNR & SSIM & LPIPS & PSNR & SSIM & LPIPS & PSNR & SSIM & LPIPS & PSNR & SSIM & LPIPS & PSNR & SSIM & LPIPS \\

     \midrule[1pt]

    Micro-Chair & 27.37 & 0.926 & 0.074 & 27.29 & 0.926 & 0.075 & 26.12 & 0.918 & 0.082 & 28.42 & 0.941 & 0.056 & \second{29.55} & \second{0.947} & \second{0.054} & \best{30.61} & \best{0.962} & \best{0.038} \\
    Micro-Ficus & 24.00 & 0.862 & 0.132 & 24.00 & 0.863 & 0.132 & 24.63 & 0.858 & 0.128 & 24.66 & \second{0.887} & \second{0.105} & \second{24.68} & 0.882 & 0.109 & \best{25.71} & \best{0.902} & \best{0.088} \\
    Grain & 33.40 & 0.925 & 0.119 & 33.42 & 0.925 & 0.119 & 33.04 & 0.925 & 0.123  & 34.19 & 0.937 & \second{0.098} & \best{34.92} & \second{0.938} & 0.099 & \second{34.77} & \best{0.942} & \best{0.086} \\
    Micro-Lego & 24.49 & 0.841 & 0.176 & 24.50 & 0.842 & 0.176 & 24.32 & 0.832 & 0.182  & 25.96 & 0.876 & 0.139 & \best{26.58} & \second{0.880} & \second{0.138} & \second{26.33} & \best{0.893} & \best{0.119} \\
    Locust & 37.60 & 0.978 & 0.036 & 37.58 & 0.978 & 0.037 & 35.57 & 0.968 & 0.051  & \best{38.86} & \best{0.983} & \second{0.026} & \second{38.64} & \second{0.982} & 0.029 & 37.92 & 0.981 & \best{0.023} \\
    Seed & \second{38.73} & 0.956 & 0.090 & 38.56 & 0.953 & 0.092 & 35.88 & 0.942 & 0.110 & 38.19 & \second{0.958} & \second{0.078} & 38.62 & 0.955 & 0.085 & \best{40.26} & \best{0.974} & \best{0.056} \\
    Micro-Ship & 25.79 & 0.818 & 0.221 & 25.78 & 0.816 & 0.222 & 25.00 & 0.801 & 0.240 & 26.20 & 0.828 & 0.212 & \second{26.33} & \second{0.831} & \second{0.207} & \best{26.57} & \best{0.842} & \best{0.171} \\
    Spider & 37.50 & 0.965 & 0.058 & 37.51 & 0.965 & 0.058 & 35.67 & 0.954 & 0.071 & \best{38.41} & \best{0.975} & \second{0.041} & 37.74 & 0.971 & 0.048 & \second{37.86} & \second{0.972} & \best{0.038} \\

    \midrule[1pt]
    
    Average & 31.11 & 0.909 & 0.113 & 31.08 & 0.909 & 0.113 & 30.03 & 0.900 & 0.123 & 31.86 & 0.923 & \second{0.094} & \second{32.13} & \second{0.923} & 0.096 & \best{32.50} & \best{0.934} & \best{0.077} \\
    
    \end{tabular}
}    
    \caption{\textbf{Quantitative comparisons on the synthetic dataset.} All metrics are averaged over the entire test set. We color code the \best{\textbf{best}} and \second{\textbf{second best}}.}
    \label{tab:results_blur}

\vspace{-1.5em}
\end{table*}

\vspace{-3mm}

\subsubsection{Multi-stage training.}

Our training is divided into three stages: initialization training, low-scale training, and high-scale training. The purpose of initialization training (first 3,000 iterations) is to create a good initial representation of the 3D scene \(\mathcal{S}\). This stage follows the standard 3DGS training procedure. The training loss \(\mathcal{L}_{\text{pre}}\) during this stage, aims to reconstruct a plausible 3D scene by directly comparing the rendered image \(\mathbf{I}_{\text{out}}\) with the input blurry image \(\mathbf{I}\):
\begin{equation}
  \mathcal{L}_{\text{pre}} = (1 - \lambda)\|\mathbf{I}-\mathbf{I}_{\text{out}}\| + \lambda \mathcal{L}_{\text{D-SSIM}}(\mathbf{I},\mathbf{I}_{\text{out}}),
  \label{eq13}
\end{equation}
where \(\mathcal{L}_{\text{D-SSIM}}\) represents the structural similarity loss. This initial supervision, while using blurry targets, helps establish a foundational geometry.

In the subsequent low-scale training stage (iterations 3,000 to 15,000), we downsample the training images \(\mathbf{I}\) (e.g., with a scaling factor \(s=2.0\)) and introduce BlurNet for joint optimization with the 3D scene. This is followed by the high-scale training stage (iterations 15,000 to 30,000) using the original image resolution (\(s=1.0\)), where a new hidden layer may be added to BlurNet to enhance its fitting capability. During both low-scale and high-scale stages, the primary training loss \(\mathcal{L}\) is employed. This loss supervises the simulated blurred image \(\widetilde{\mathbf{I}}_{\text{out}}\) generated by convolving the rendered sharp \(\mathbf{I}_{\text{out}}\) with predicted kernels by BlurNet to match the input blurry image \(\mathbf{I}\), while also applying a total variation regularization to the rendered sharp image \(\mathbf{I}_{\text{out}}\) to encourage smoothness:
\begin{equation}
  \mathcal{L} = (1 - \lambda)\|\mathbf{I}-\widetilde{\mathbf{I}}_{\text{out}}\| + \lambda \mathcal{L}_{\text{D-SSIM}}(\mathbf{I}, \widetilde{\mathbf{I}}_{\text{out}}) + \mathcal{L}_{\text{TV}}(\mathbf{I}_{\text{out}}).
  \label{eq14}
\end{equation}


By replicating the physical blurring process in rendering with priors from macrophotography, the end-to-end framework inherently constrains the solution space. This structural regularization enables effective supervision.

This multi-stage training strategy has several advantages: any degree of blurry pixels can be effectively modeled in multi-scale training, without multi-scale training, some kinds of blur kernels cannot be modeled; low-scale BlurNet provides an effective warm-up for high-scale training; and since half of the iterations are performed at low scales, training time is reduced by about half compared to using only high-scale inputs. 

\section{Experiments}
\label{sec:experiments}

\subsection{Implementation Details}

\subsubsection{Training Settings.}
We use the Adam optimizer~\cite{kingma2017adam} and set the learning rate of the two CNNs \(\mathcal{A}\) and \(\mathcal{E}\) to 0.001. For 2D Gaussian kernel, we set the pixel kernel size of each pixel to \(2K+1=13\), \ie \(K=6\), and the other hyperparameters are the same as those in 3D-GS. \(\mathcal{A}\) and \(\mathcal{E}\) are both 2-layer CNNs with 32 hidden units in each layer and ReLU as the activation function. Only the first layer is applied in the low-scale training stage, while the second is added during high-scale training. The frequency \(L\) in positional encoding \(P_L(\cdot)\) is 7. The total number of training iterations is 30,000, of which the first 3,000 iterations are pre-training, 3,000 to 15,000 iterations are low-scale training, and 15,000 to 30,000 iterations are high-scale training. All experiments are performed on an NVIDIA RTX 4090 GPU.

\vspace{-3mm}
\subsubsection{Datasets.}
We use a camera with a macro lens to take multi-view images of 5 common subcentimeter tiny objects as a real dataset for evaluation, each of which contains about 200 images. The macro lens we use is an Olympus M.Zuiko Digital ED 60mm F2.8 Macro, paired with a Panasonic Lumix DC-GH5 camera. We use a turnable device to capture multi-view images, which enables us to change object pose easily while these objects are too small to change poses. Considering the difficulty for COLMAP~\cite{misc_colmap1, misc_colmap2} to estimate camera poses under extreme camera parameters and critical defocus blur, we use a more professional software Agisoft Metashape~\cite{AgiSoftPhotoScan} to calculate the camera poses of blurry images and reference images in these real scenes, and then convert them to the form that COLMAP can read to check correctness. In addition, we selected 4 objects from the NeRF dataset~\cite{mildenhall2020nerf} and selected 4 common small objects, which were carefully processed into macro-scale models using Blender\cite{misc_blender} as the synthetic dataset, and rendered 100 multi-view images for each object with reference to real-world camera parameters. It contains both blurry images and clear images from the same view, which is convenient for calculating quantitative results. More dataset settings can be found in the Supplementary Material.


\subsection{Comparisons}

Due to significant differences between macrophotography and natural imaging, existing natural deblurring methods~\cite{Zhang_2024_CVPR_mptcatablur, zhang_benchmarking_2022} are not suitable for macrophotography deblurring. To the best of our knowledge, no studies have attempted to reconstruct tiny objects from multi-view blurred macrophotography inputs. We compare with several potentially effective methods: (1) 3DGS-based deblurring methods including BAGS~\cite{peng2024bags}, Deblurring-3DGS~\cite{lee2024deblurring3dgs}, and naive 3DGS~\cite{kerbl20233dgs}; (2) image-space deblurring methods that perform well in macro scenes, combined with 3DGS. We evaluated LaKDNet + 3DGS~\cite{ruan2023}, Restormer-TLC + 3DGS~\cite{10.1007/978-3-031-20071-7_4}, Restormer + 3DGS~\cite{Zamir2021Restormer}, and INIKNet + 3DGS~\cite{quan2023}, and selected the two best-performing methods~\cite{Zamir2021Restormer, quan2023} for comparison. Following 3DGS, we use PSNR, SSIM, and LPIPS as evaluation metrics for novel view synthesis quality.

\begin{figure*}[htb!]
    \setlength{\abovecaptionskip}{3pt}
    \setlength{\tabcolsep}{0.5pt}
    \centering
    \small
    \begin{tabular}[b]{ccccccccc}
         Novel View & Defocus & 3DGS\cite{kerbl20233dgs} & DbGS\cite{lee2024deblurring3dgs} & BAGS\cite{peng2024bags} & Res\cite{Zamir2021Restormer} & INIK\cite{quan2023} & Ours & GT\\

        \begin{subfigure}[b]{0.18\linewidth}
            \includegraphics[width=\textwidth]{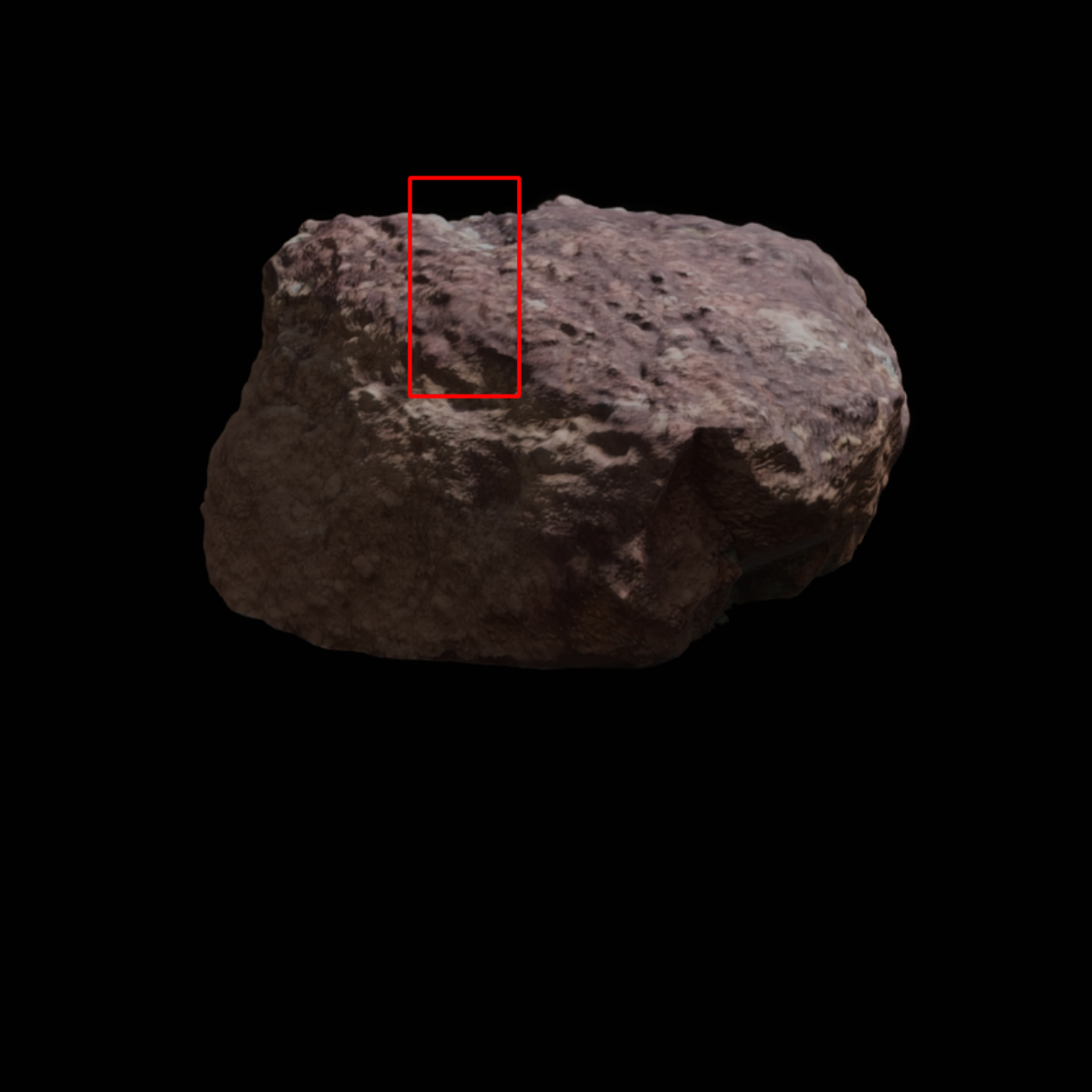}
        \end{subfigure} &
        \begin{subfigure}[b]{0.09\linewidth}
            \includegraphics[width=\textwidth]{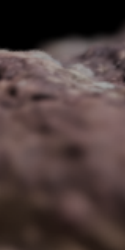}
        \end{subfigure} &  
        \begin{subfigure}[b]{0.09\linewidth}
            \includegraphics[width=\textwidth]{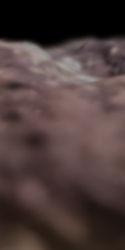}
        \end{subfigure} &
        \begin{subfigure}[b]{0.09\linewidth}
            \includegraphics[width=\textwidth]{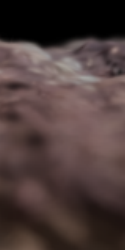}
        \end{subfigure} &
        \begin{subfigure}[b]{0.09\linewidth}
            \includegraphics[width=\textwidth]{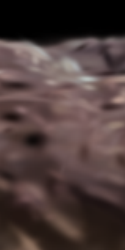}
        \end{subfigure} &
        \begin{subfigure}[b]{0.09\linewidth}
            \includegraphics[width=\textwidth]{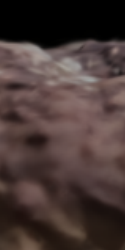}
        \end{subfigure} &
        \begin{subfigure}[b]{0.09\linewidth}
            \includegraphics[width=\textwidth]{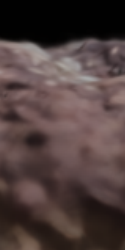}
        \end{subfigure} &   
        \begin{subfigure}[b]{0.09\linewidth}
            \includegraphics[width=\textwidth]{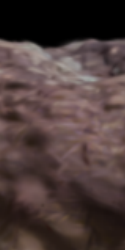}
        \end{subfigure} &  
        \begin{subfigure}[b]{0.09\linewidth}
            \includegraphics[width=\textwidth]{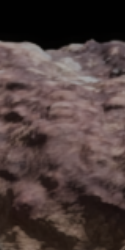}
        \end{subfigure} \\ 

        PSNR/SSIM &  & 33.40/0.925 & 33.42/0.925 & 33.04/0.925 & 34.19/0.937 & \best{34.92}/\second{0.938} & \second{34.77}/\best{0.942} & \\
        
        \begin{subfigure}[b]{0.18\linewidth}
            \includegraphics[width=\textwidth]{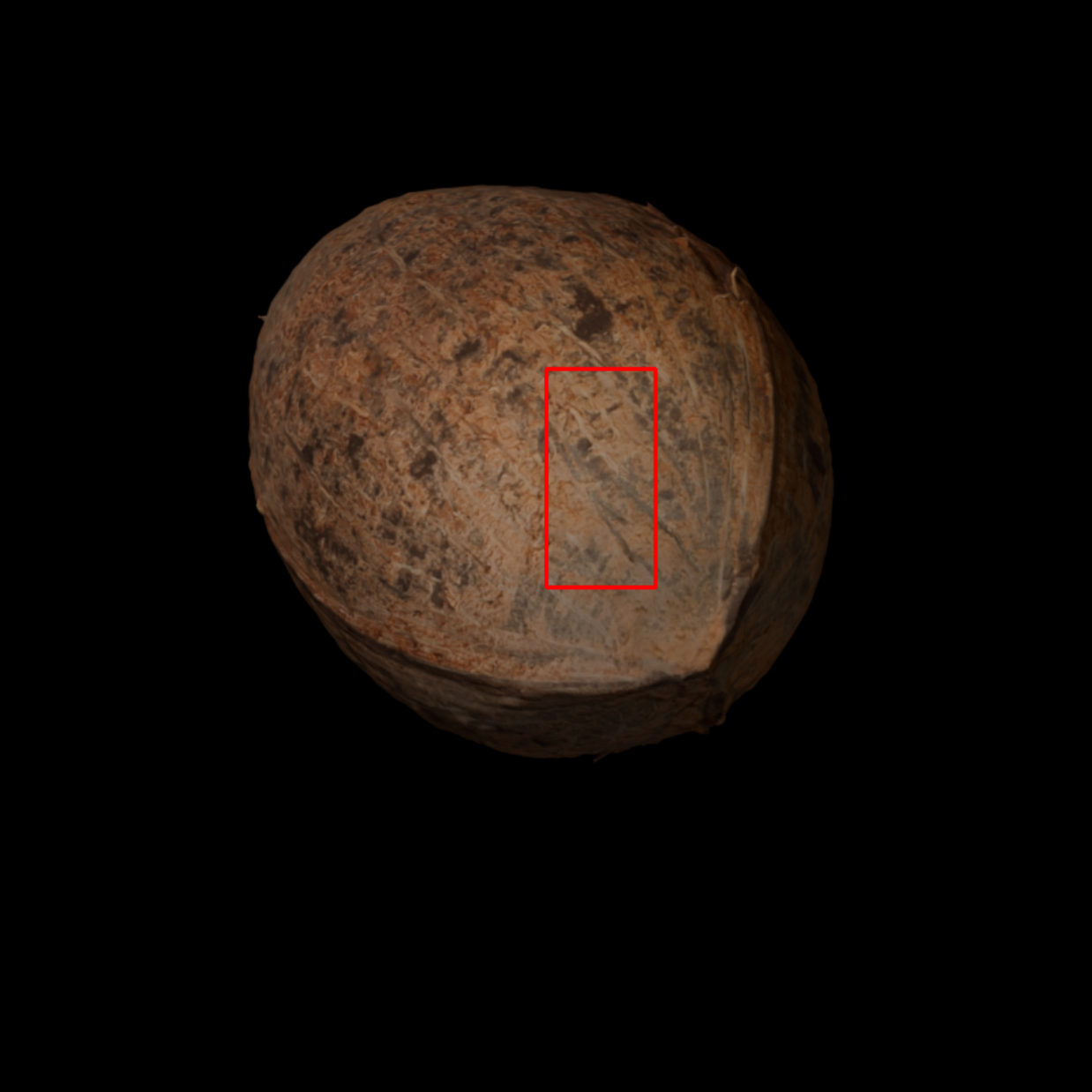}
        \end{subfigure} &
        \begin{subfigure}[b]{0.09\linewidth}
            \includegraphics[width=\textwidth]{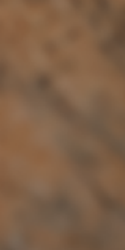}
        \end{subfigure} &  
        \begin{subfigure}[b]{0.09\linewidth}
            \includegraphics[width=\textwidth]{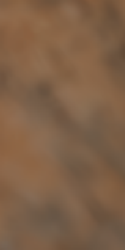}
        \end{subfigure} &
        \begin{subfigure}[b]{0.09\linewidth}
            \includegraphics[width=\textwidth]{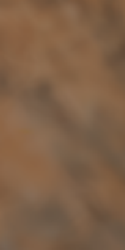}
        \end{subfigure} &
        \begin{subfigure}[b]{0.09\linewidth}
            \includegraphics[width=\textwidth]{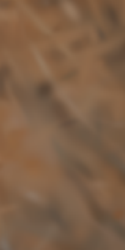}
        \end{subfigure} &
        \begin{subfigure}[b]{0.09\linewidth}
            \includegraphics[width=\textwidth]{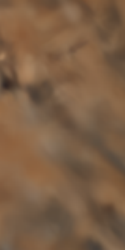}
        \end{subfigure} &
        \begin{subfigure}[b]{0.09\linewidth}
            \includegraphics[width=\textwidth]{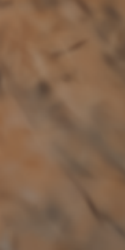}
        \end{subfigure} &   
        \begin{subfigure}[b]{0.09\linewidth}
            \includegraphics[width=\textwidth]{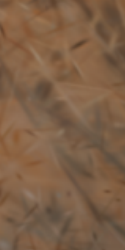}
        \end{subfigure} &  
        \begin{subfigure}[b]{0.09\linewidth}
            \includegraphics[width=\textwidth]{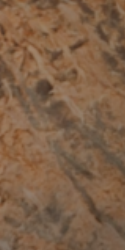}
        \end{subfigure} \\

        PSNR/SSIM &  & \second{38.73}/0.956 & 38.56/0.953 & 35.88/0.942 & 38.19/\second{0.958} & 38.62/0.955 & \best{40.186}/\best{0.974} & \\

        \begin{subfigure}[b]{0.18\linewidth}
            \includegraphics[width=\textwidth]{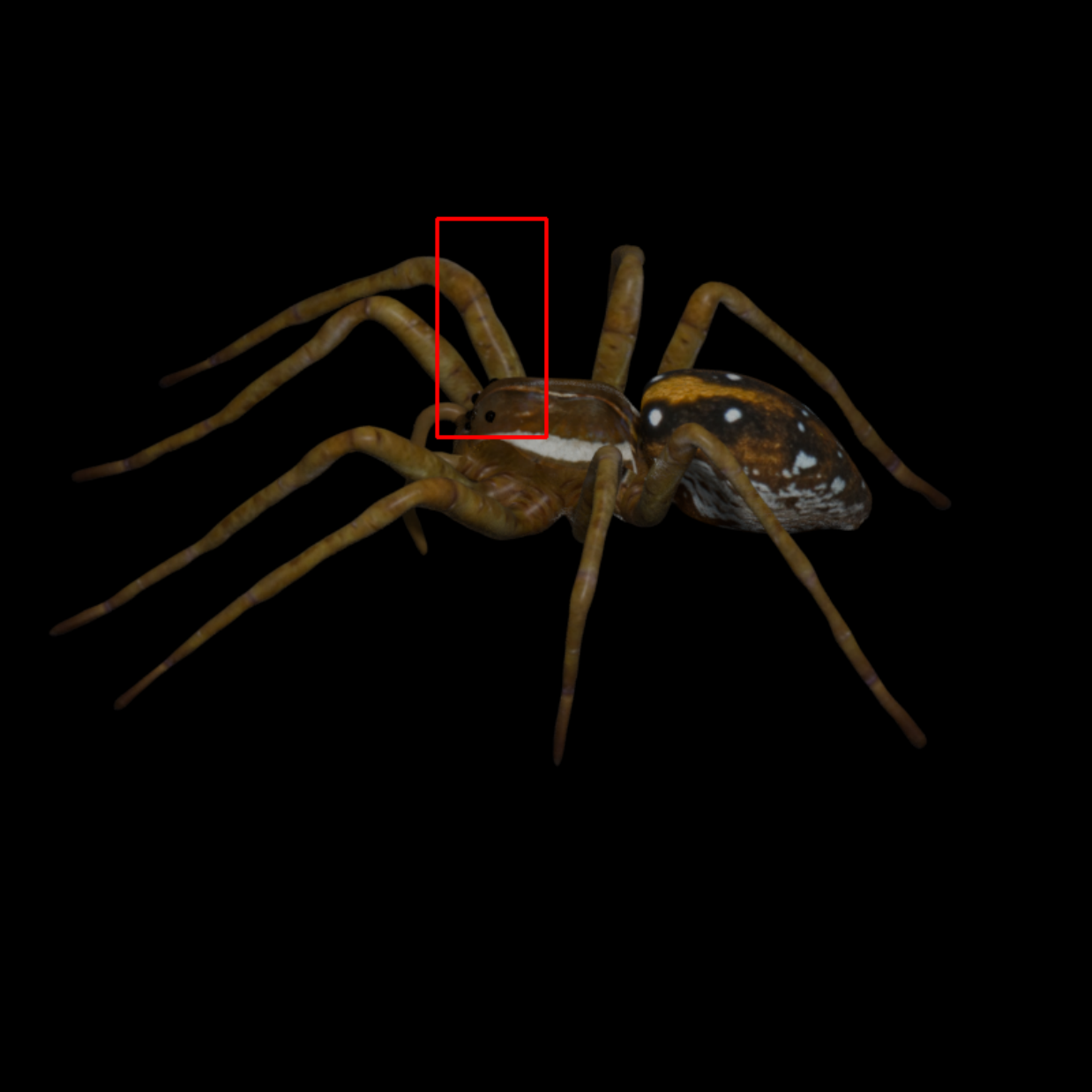}
        \end{subfigure} &
        \begin{subfigure}[b]{0.09\linewidth}
            \includegraphics[width=\textwidth]{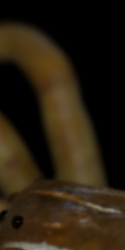}
        \end{subfigure} &  
        \begin{subfigure}[b]{0.09\linewidth}
            \includegraphics[width=\textwidth]{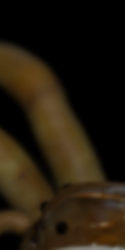}
        \end{subfigure} &
        \begin{subfigure}[b]{0.09\linewidth}
            \includegraphics[width=\textwidth]{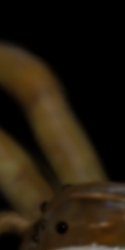}
        \end{subfigure} &
        \begin{subfigure}[b]{0.09\linewidth}
            \includegraphics[width=\textwidth]{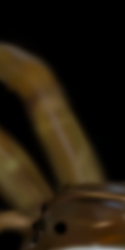}
        \end{subfigure} &
        \begin{subfigure}[b]{0.09\linewidth}
            \includegraphics[width=\textwidth]{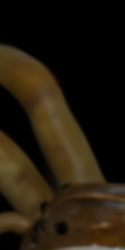}
        \end{subfigure} &
        \begin{subfigure}[b]{0.09\linewidth}
            \includegraphics[width=\textwidth]{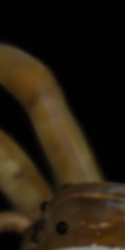}
        \end{subfigure} &   
        \begin{subfigure}[b]{0.09\linewidth}
            \includegraphics[width=\textwidth]{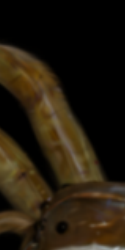}
        \end{subfigure} &  
        \begin{subfigure}[b]{0.09\linewidth}
            \includegraphics[width=\textwidth]{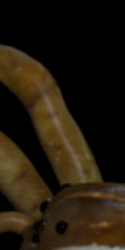}
        \end{subfigure} \\ 

        PSNR/SSIM &  & 37.50/0.965 & 37.51/0.965 & 35.67/0.954 & \best{38.41}/\best{0.975} & 37.74/0.971 & \second{37.86}/\second{0.972} & \\

    \end{tabular}
    \caption{\textbf{Visual comparisons on novel views of synthetic images dataset}. DbGS, Res, and INIK are the shorthands for Deblurring-3DGS, Restormer+3DGS, and INIKNet+3DGS. We color code the \best{\textbf{best}} and the \second{\textbf{second best}}.}
    \label{fig:syn_comparison}
\end{figure*}

\begin{figure*}[htb!]
    \vspace{-0.5em}
    \setlength{\abovecaptionskip}{3pt}
    \setlength{\tabcolsep}{0.5pt}
    \centering
    \small
    \begin{tabular}[b]{ccccccccc}
         Defocused Input & Defocus & 3DGS\cite{kerbl20233dgs} & DbGS\cite{lee2024deblurring3dgs} & BAGS\cite{peng2024bags} & Res\cite{Zamir2021Restormer} & INIK\cite{quan2023} & Ours & Reference\\

        \begin{subfigure}[b]{0.18\linewidth}
            \includegraphics[width=\textwidth]{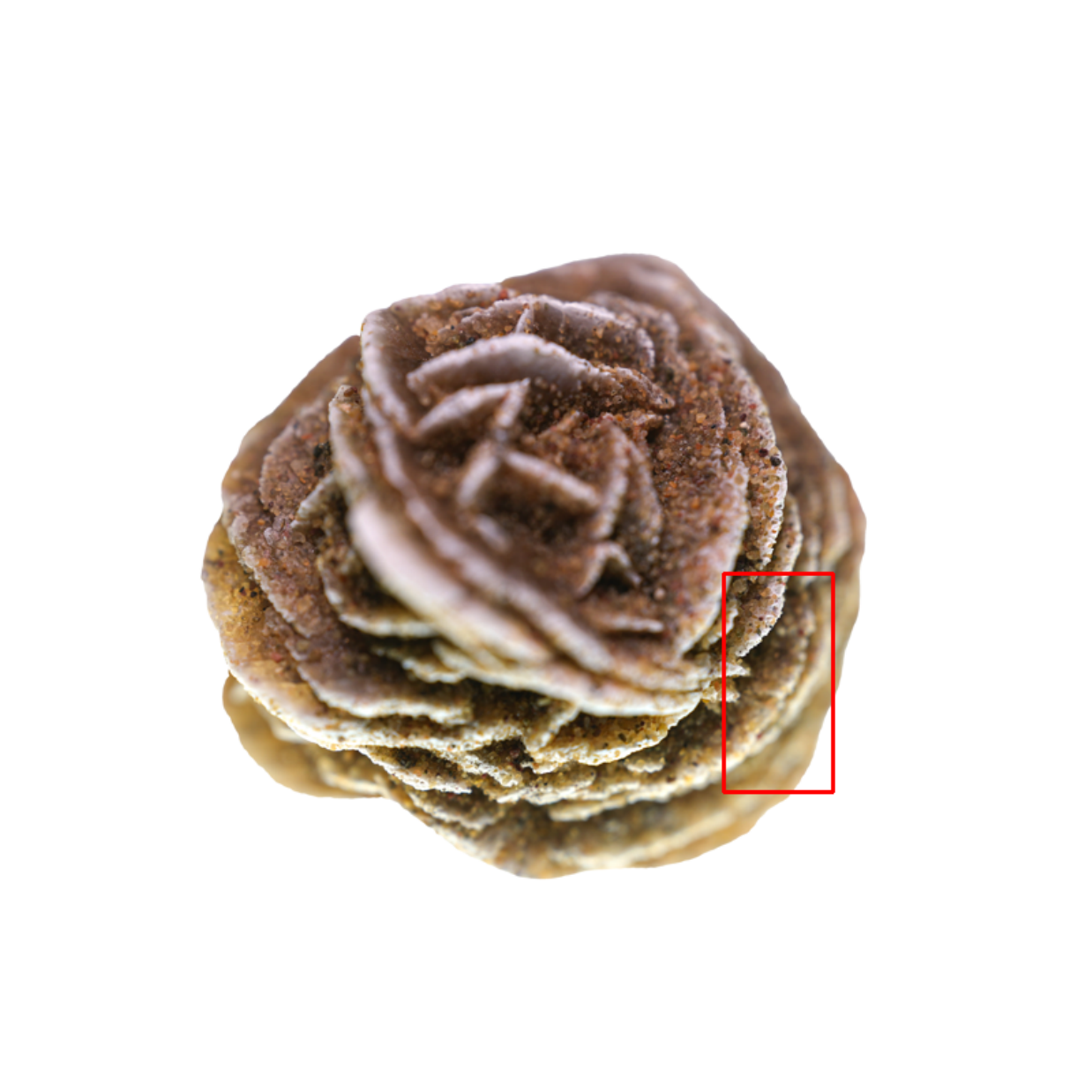}
        \end{subfigure} &
        \begin{subfigure}[b]{0.09\linewidth}
            \includegraphics[width=\textwidth]{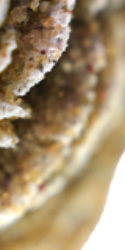}
        \end{subfigure} &  
        \begin{subfigure}[b]{0.09\linewidth}
            \includegraphics[width=\textwidth]{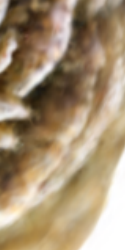}
        \end{subfigure} &
        \begin{subfigure}[b]{0.09\linewidth}
            \includegraphics[width=\textwidth]{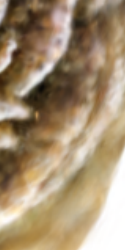}
        \end{subfigure} &
        \begin{subfigure}[b]{0.09\linewidth}
            \includegraphics[width=\textwidth]{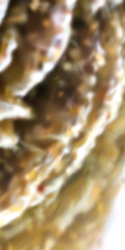}
        \end{subfigure} &
        \begin{subfigure}[b]{0.09\linewidth}
            \includegraphics[width=\textwidth]{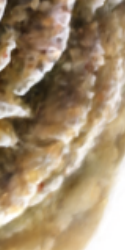}
        \end{subfigure} &
        \begin{subfigure}[b]{0.09\linewidth}
            \includegraphics[width=\textwidth]{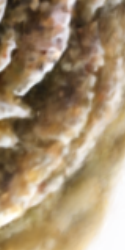}
        \end{subfigure} &   
        \begin{subfigure}[b]{0.09\linewidth}
            \includegraphics[width=\textwidth]{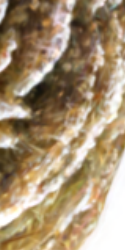}
        \end{subfigure} &  
        \begin{subfigure}[b]{0.09\linewidth}
            \includegraphics[width=\textwidth]{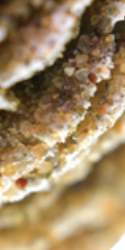}
        \end{subfigure} \\ 


        \begin{subfigure}[b]{0.18\linewidth}
            \includegraphics[width=\textwidth]{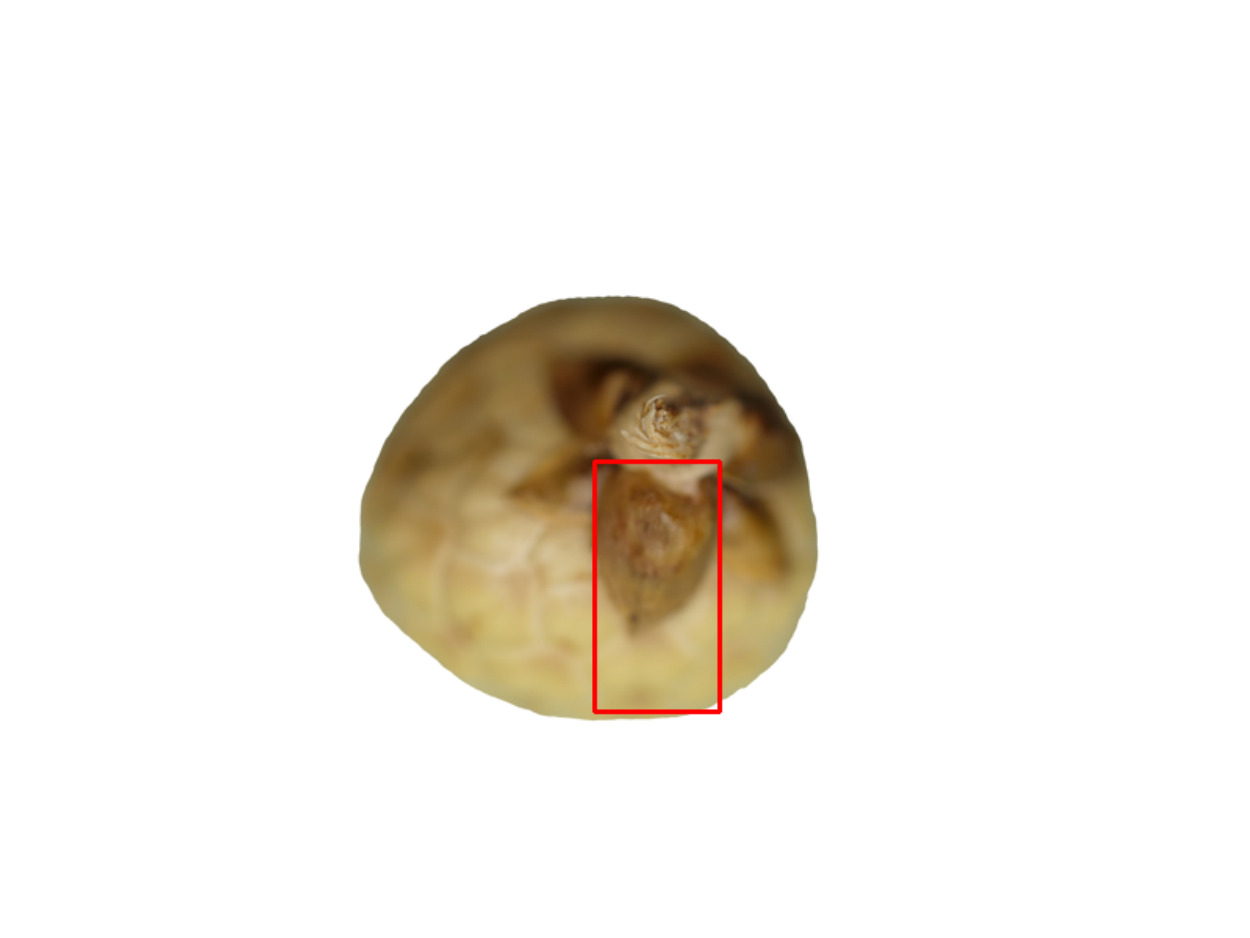}
        \end{subfigure} &
        \begin{subfigure}[b]{0.09\linewidth}
            \includegraphics[width=\textwidth]{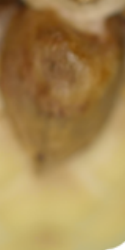}
        \end{subfigure} &  
        \begin{subfigure}[b]{0.09\linewidth}
            \includegraphics[width=\textwidth]{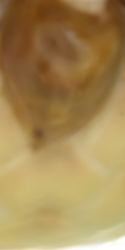}
        \end{subfigure} &
        \begin{subfigure}[b]{0.09\linewidth}
            \includegraphics[width=\textwidth]{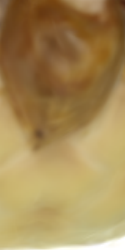}
        \end{subfigure} &
        \begin{subfigure}[b]{0.09\linewidth}
            \includegraphics[width=\textwidth]{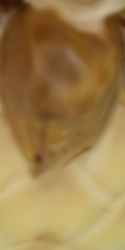}
        \end{subfigure} &
        \begin{subfigure}[b]{0.09\linewidth}
            \includegraphics[width=\textwidth]{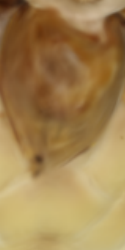}
        \end{subfigure} &
        \begin{subfigure}[b]{0.09\linewidth}
            \includegraphics[width=\textwidth]{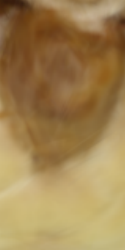}
        \end{subfigure} &   
        \begin{subfigure}[b]{0.09\linewidth}
            \includegraphics[width=\textwidth]{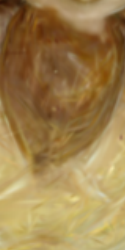}
        \end{subfigure} &  
        \begin{subfigure}[b]{0.09\linewidth}
            \includegraphics[width=\textwidth]{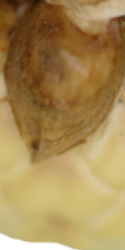}
        \end{subfigure} \\ 


        \begin{subfigure}[b]{0.18\linewidth}
            \includegraphics[width=\textwidth]{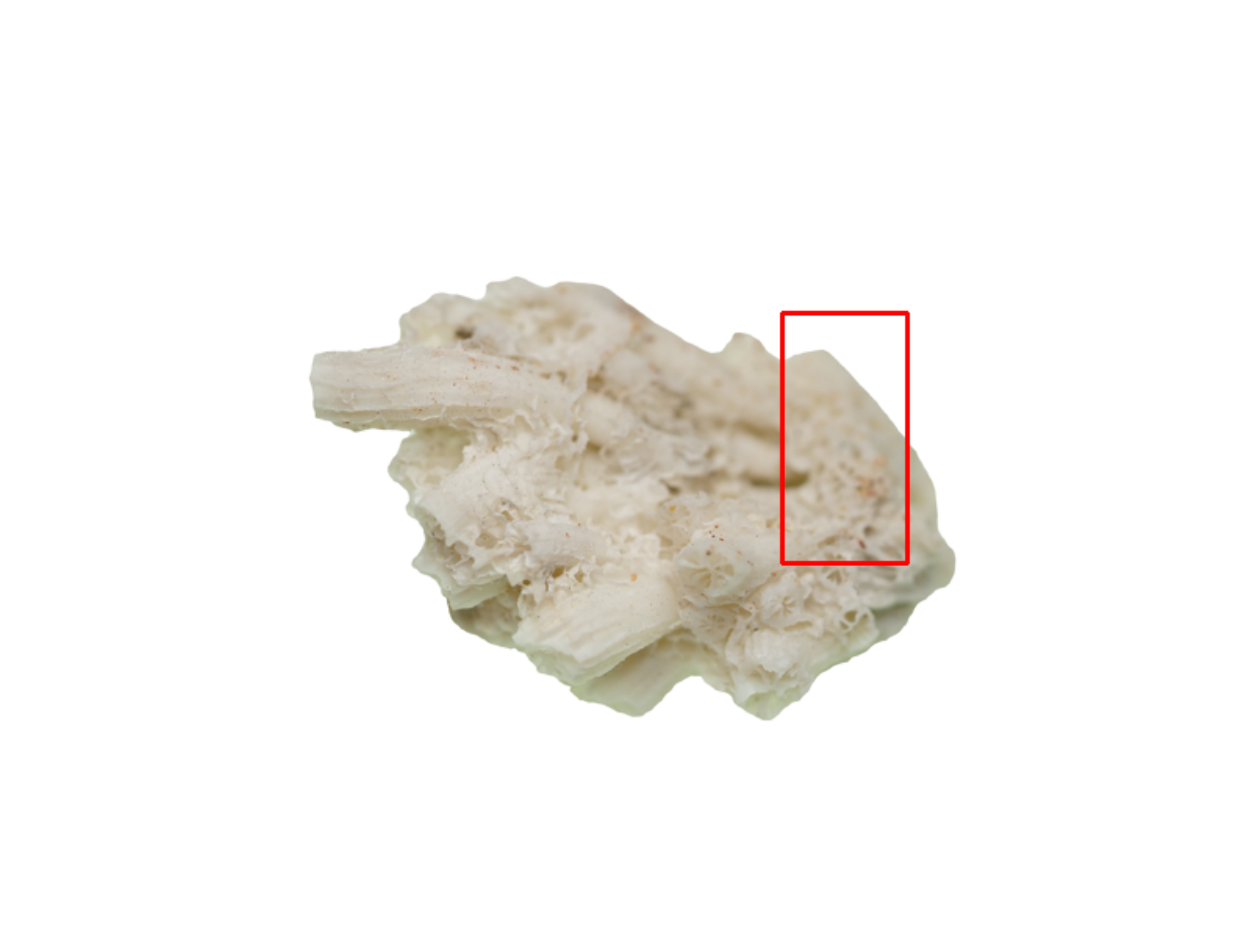}
        \end{subfigure} &
        \begin{subfigure}[b]{0.09\linewidth}
            \includegraphics[width=\textwidth]{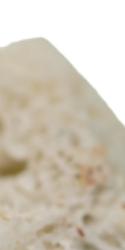}
        \end{subfigure} &  
        \begin{subfigure}[b]{0.09\linewidth}
            \includegraphics[width=\textwidth]{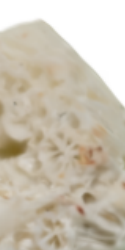}
        \end{subfigure} &
        \begin{subfigure}[b]{0.09\linewidth}
            \includegraphics[width=\textwidth]{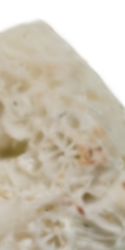}
        \end{subfigure} &
        \begin{subfigure}[b]{0.09\linewidth}
            \includegraphics[width=\textwidth]{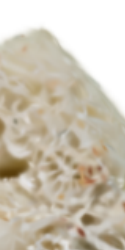}
        \end{subfigure} &
        \begin{subfigure}[b]{0.09\linewidth}
            \includegraphics[width=\textwidth]{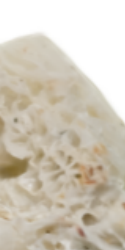}
        \end{subfigure} &
        \begin{subfigure}[b]{0.09\linewidth}
            \includegraphics[width=\textwidth]{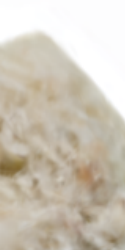}
        \end{subfigure} &   
        \begin{subfigure}[b]{0.09\linewidth}
            \includegraphics[width=\textwidth]{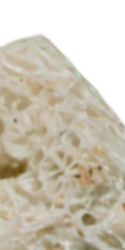}
        \end{subfigure} &  
        \begin{subfigure}[b]{0.09\linewidth}
            \includegraphics[width=\textwidth]{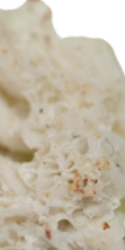}
        \end{subfigure} \\ 


    \end{tabular}
    \caption{\textbf{Visual comparisons on test views real-world macrophotography images dataset.} DbGS, Res, and INIK are the shorthands for Deblurring-3DGS, Restormer+3DGS, and INIKNet+3DGS. Macro lenses can not capture all-in-focus images. Therefore, we can only compare our results with clear parts in the reference images. For the same reason, quantitative comparison cannot be carried out. }
    \label{fig:real_comparison}
    \vspace{-1.5em}
\end{figure*}

Tab.~\ref{tab:results_blur} shows the quantitative comparison results. As shown in Tab.~\ref{tab:results_blur}, our method performs better than other advanced deblurring and reconstruction methods, especially achieving significant improvements on SSIM and LPIPS. This shows that our method maintains good geometric characteristics during deblurring and reconstruction, keeping multi-view consistency. Note that BAGS and Deblurring-3DGS perform slightly worse than naive 3DGS. One possible reason is that their approaches introduce depth but handle it improperly. The importance and sensitivity of depth features in macrophotography lead to their performance degradation. 
Fig.~\ref{fig:real_comparison} and Fig.~\ref{fig:syn_comparison} show the qualitative comparison results on synthetic and captured real datasets. The images rendered by our method have sharp edges and rich details, which are closest to reality. Other methods suffer from artifacts near object boundaries and blurred textures. For more reconstruction results, please see our demo video.

\subsection{Ablation Studies}

\subsubsection{Evaluation of the blur kernel.}
\begin{figure}[htb!]
    \setlength{\abovecaptionskip}{3pt}
    \setlength{\tabcolsep}{1pt}
    \centering
    \small
    \begin{tabular}[b]{cccc}
        
        \begin{subfigure}[b]{0.25\linewidth}
            \includegraphics[width=\textwidth]{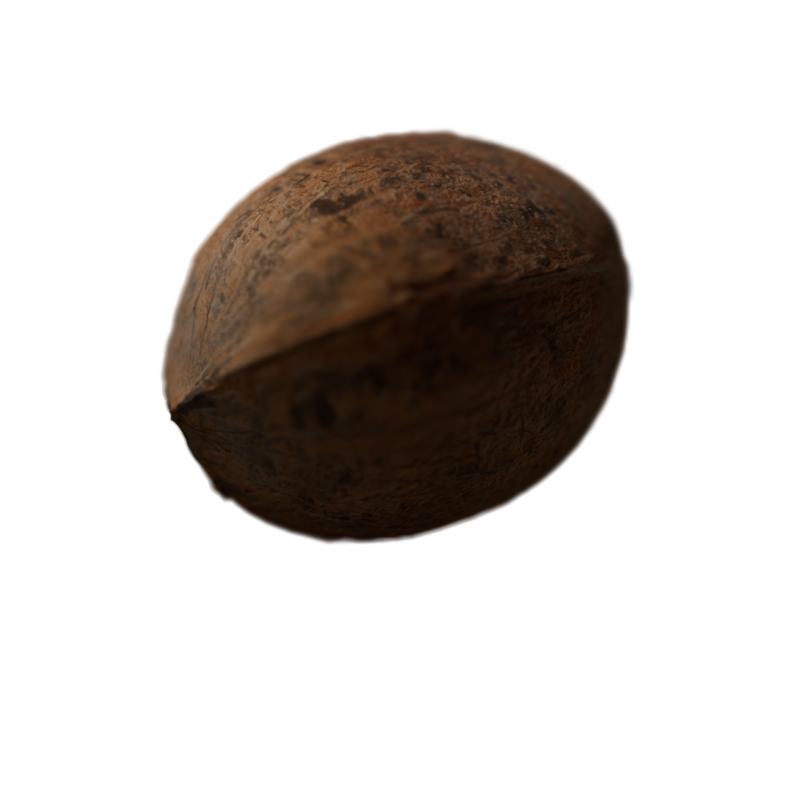}
        \end{subfigure} &
        \begin{subfigure}[b]{0.25\linewidth}
            \includegraphics[width=\textwidth]{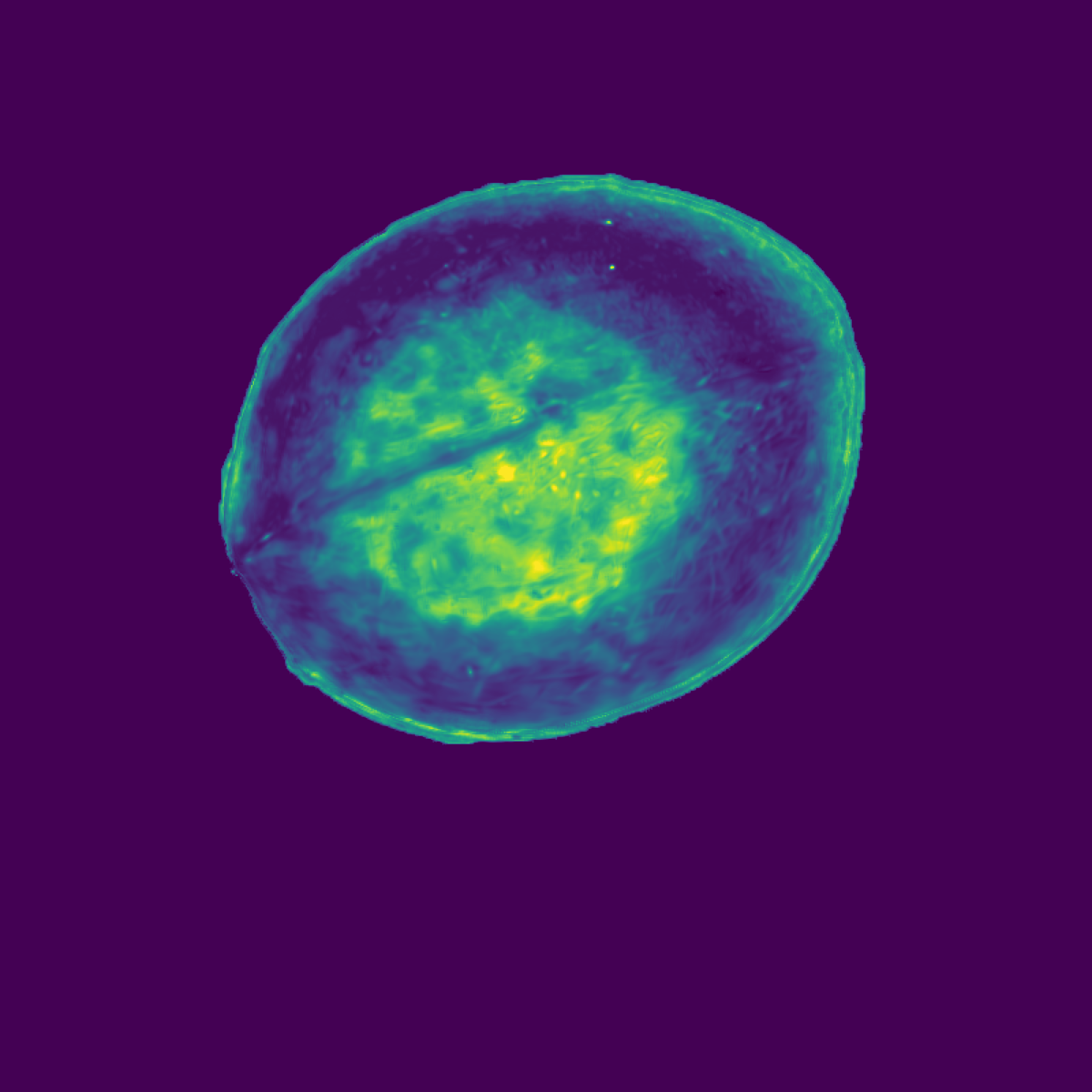}
        \end{subfigure} &
        \begin{subfigure}[b]{0.25\linewidth}
            \includegraphics[width=\textwidth]{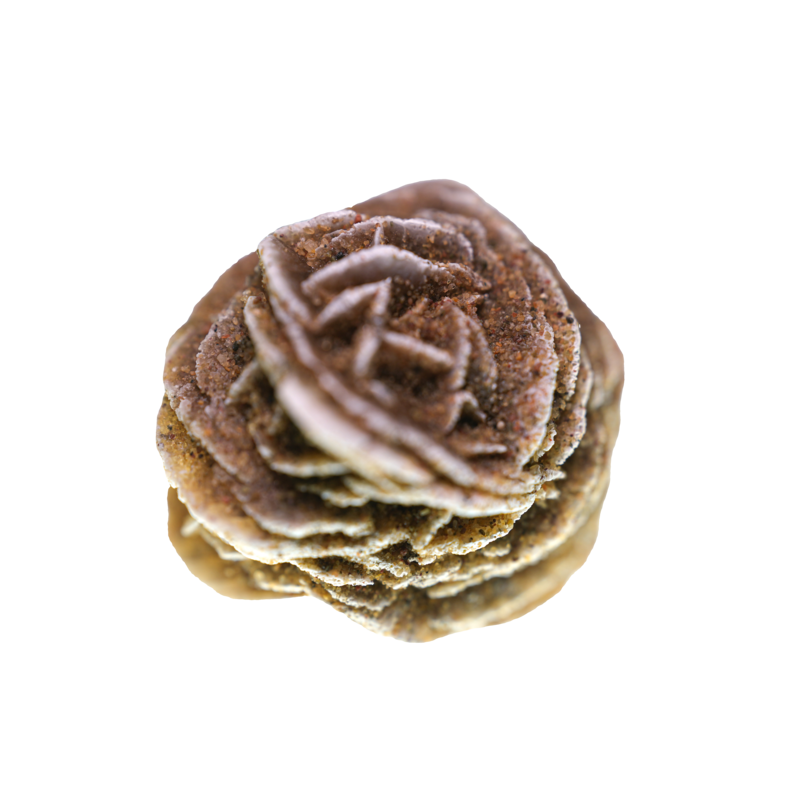}
        \end{subfigure} &
        \begin{subfigure}[b]{0.25\linewidth}
            \includegraphics[width=\textwidth]{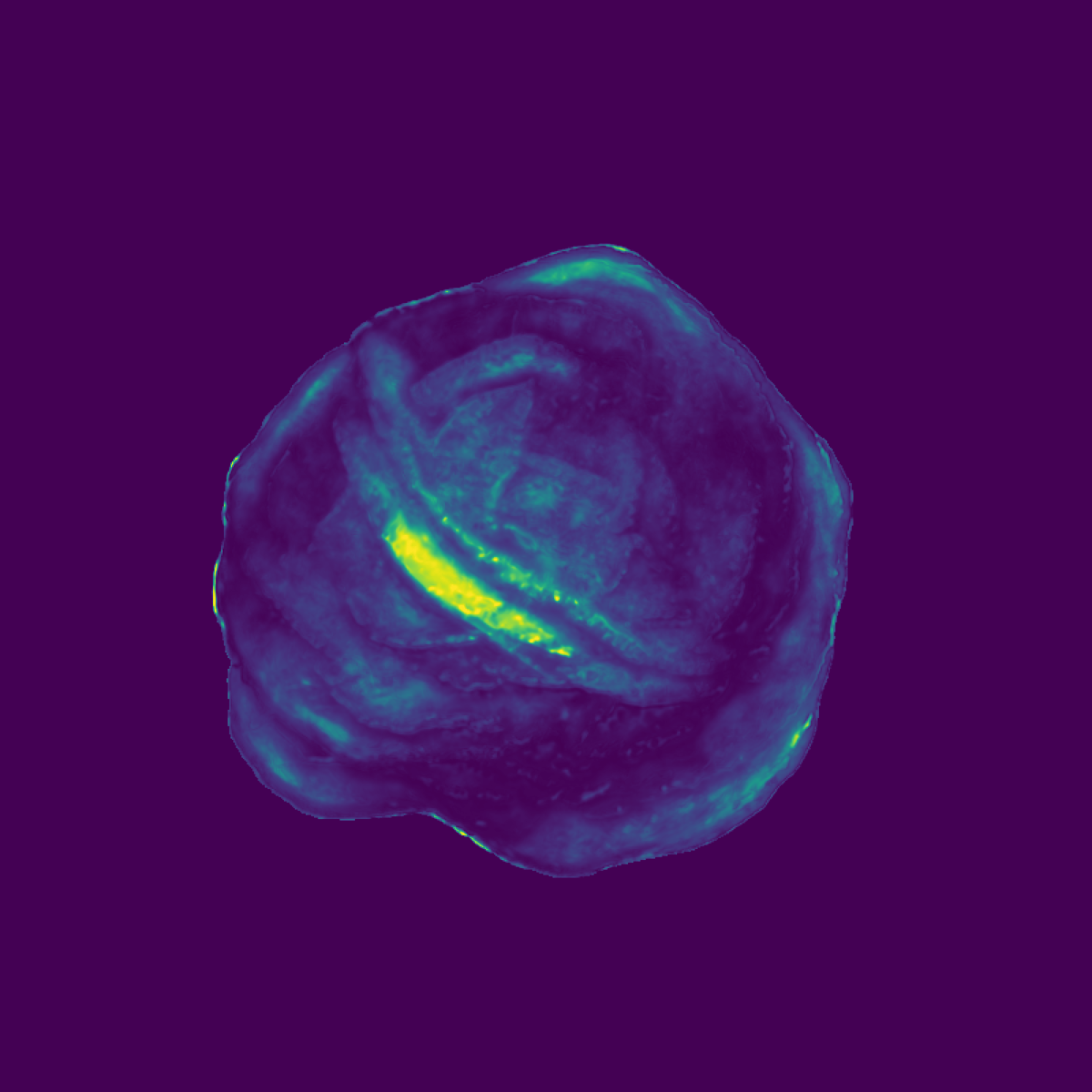}
        \end{subfigure} \\
        Defocus & Variance Map & Defocus & Variance Map \\
        
        \multicolumn{2}{c}{Synthetic Data} & \multicolumn{2}{c}{Real Data} \\
        
    \end{tabular}
    \vspace{-1em}
    \caption{\textbf{Visualization of estimated variance maps}. Comparison between defocused input images and estimated variance maps qualitatively demonstrates our method's ability to accurately predict larger blur kernel variances in defocused regions.}
    \label{fig:blur_visualization}
\end{figure}

We validate the effectiveness of our blur kernels from two perspectives. First, Fig.~\ref{fig:blur_visualization} demonstrates that our BlurNet accurately estimates variance maps that correspond well to the defocused regions in the input images, showing larger variance values in more blurred areas for both synthetic and real macrophotography data. Then, we quantitatively evaluate the blur kernel reconstruction accuracy on synthetic data. Clean images generated by Blender are used to construct their corresponding ground truth 3DGS models. The clean images rendered from novel views are then reblurred using the blur kernels recovered by our method and compared with the real blurred images from the corresponding views. Tab.~\ref{tab:rebuttal_comp} shows that our blur kernels can accurately reconstruct the defocus blur in the blurred images.

\begin{table}[htbp]
\vspace{-0.5em}
    \centering
    \resizebox{\linewidth}{!}{

    \begin{tabular}{ l | cccc | c}

      Metric & Grain & Locust & Seed & Spider & Average \\

     \midrule[1pt]

     PSNR \(\uparrow\) & 33.58/41.35 & 37.98/41.92 & 39.86/45.69 & 37.54/41.35 & 37.24/42.58 \\ 
     SSIM \(\uparrow\) & 0.931/0.989 & 0.980/0.991 & 0.967/0.991 & 0.968/0.986 & 0.962/0.989 \\
     LPIPS \(\downarrow\) & 0.108/0.037 & 0.033/0.017 & 0.077/0.031 & 0.050/0.027 & 0.067/0.028 \\

    \end{tabular}
}    
    \vspace{-1em}
    \caption{\textbf{Quantitative evaluation of blur kernel accuracy.} Left: Metrics between the clear renders and GT blurry images. Right: Metrics between our re-blurred results and GT blurry images.}
    \label{tab:rebuttal_comp}

\end{table}

\vspace{-1.0em}
\subsubsection{Effectiveness of Components.}
We conduct ablation studies on the four components of our method to demonstrate their effectiveness: 

\begin{itemize}
    \item \textbf{Multi-Stage Training}: We compare the results of multi-stage training with those of a single-stage training where only high-scale inputs are used.
    \item \textbf{Clarity Mask}:
    We compare our method with and without a clarity mask \(\mathbf{M}\). Without \(\mathbf{M}\), we only use the RGB image and depth map as BlurNet input.
    \item \textbf{BlurNet}: 
    We use a simple 2-layer CNN instead of BlurNet to implement the method. The clarity mask, RGB image, and depth map are concatenated as CNN input.
    \item \textbf{Depth Input}: 
    We remove the depth map generated in training process. Without depth, BlurNet can only predict the blur kernel from RGB information.
\end{itemize}

Tab.~\ref{tab:quan_ablation} shows the ablation study results for this part. Overall, the best results are achieved when using the full model. The clarity mask provides additional information about the camera focal plane in the blurry input and plays an important role in handling depth. Our carefully designed network further improves the performance of our method. Multi-stage training enables our method to model blur kernels of all sizes. Without it, kernels that are too large cannot be modeled. Given defocus blur as a physically inherent function of depth, depth information introduces physical constraints for BlurNet and plays a key role in deblurring. When various strategies cooperate, our method achieves improvements in all metrics. Detailed visualization results of ablation studies are provided in the Supplementary Material.

\subsubsection{Size of Blur Kernels.}
\begin{figure}[htb!]
    \vspace{-1em}
    \setlength{\abovecaptionskip}{3pt}
    \setlength{\tabcolsep}{0.5pt}
    \centering
    \small
    \begin{tabular}[b]{cc}

        \begin{subfigure}[b]{0.5\linewidth}
            \includegraphics[width=\textwidth]{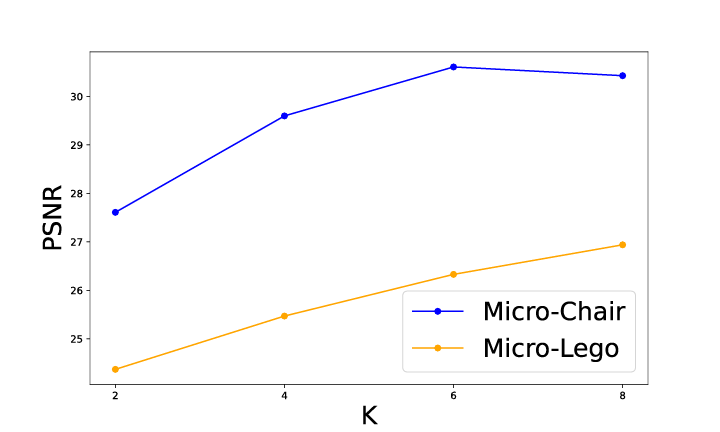}
            \caption{(a)}
            \label{fig:ablation_k_a}
        \end{subfigure} &
        \begin{subfigure}[b]{0.5\linewidth}
            \includegraphics[width=\textwidth]{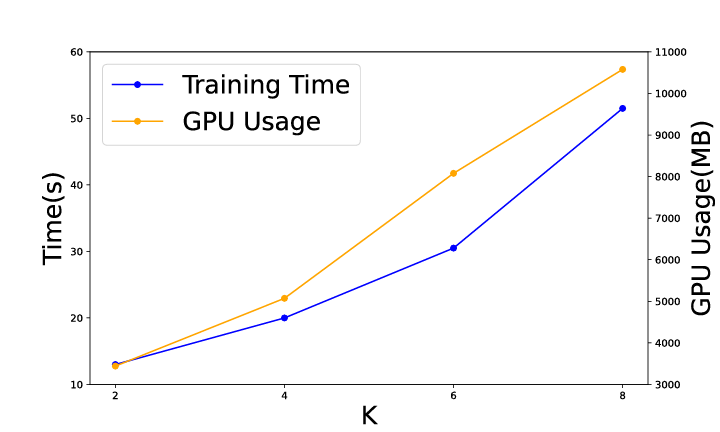}
            \caption{(b)}
            \label{fig:ablation_k_b}
        \end{subfigure}  \\ 


    \end{tabular}
    \caption{\textbf{Ablation studies on hyper-parameter kernel size \(K\)}. In (a), we measure the PSNR and SSIM of two scenes under different values of \(K\). (b) compares the training time and GPU memory usage under different values of \(K\).}
    \label{fig:ablation_k}
    \vspace{-3em}
\end{figure}

The size of the defocus blur kernel is a key hyper-parameter in our method. We study the impact of the blur kernel size $K$ on the reconstruction quality in these scenarios. We try different \(K\) values using the full model on our dataset. Fig.~\ref{fig:ablation_k_a} shows that larger kernels generally lead to higher performance. However, when we set \(K=8\) (\ie \(2K+1=17\)), the performance is almost the same as \(K=6\), while Fig.~\ref{fig:ablation_k_b} shows large kernels severely slow down training and may cause OOM issues. Considering these factors, we choose \(K=6\) as a balanced choice.

\section{Conclusion}
\label{sec:conclusion}

This paper presented a novel self-supervised method for joint defocus deblurring and 3D reconstruction, specifically tailored for macrophotography of sub-centimeter scale objects. Our approach uniquely models spatially-varying defocus blur kernels, leveraging 3D scene information like depth from 3D scene via our BlurNet module to accurately reflect macro-specific optical effects. This enables simultaneous optimization of a sharp 3D scene and the blur model using differentiable rendering, without requiring a mount of paired training data. The key insight is that the inherent multi-view consistency constraints in 3D reconstruction and optical prior in macrophotography provide sufficient supervision to jointly learn both scene geometry and blur characteristics, eliminating the need for explicit supervision. Extensive experiments demonstrate superior performance in both image deblurring and 3D reconstruction quality compared to existing methods. Future work will focus on enhancing depth estimation robustness and exploring applications in broader macro imaging contexts.






\vspace{-3em}
\section{Acknowledgement}
\label{sec:acknowledgement}

\vspace{-0.2em}

This research was supported by the National Natural Science Foundation of China(No.62441224, No.62272433), and the Fundamental Research Funds for the Central Universities.

\begin{table}[htbp]
    \vspace{-0.5em}
    \centering
    \resizebox{\linewidth}{!}{

    \begin{tabular}{l |cccc | c}


        Average & w/o stages & w/o mask & w/o BlurNet & w/o depth & Ours \\

     \midrule[1pt]

     PSNR & 28.32 & 30.95 & 31.76 & 31.56 & \best{32.50} \\
     SSIM & 0.890 & 0.908 & 0.930 & 0.918 & \best{0.934} \\
     LPIPS & 0.107 & 0.107 & 0.081 & 0.101 & \best{0.077} \\
    
    \end{tabular}
}    
    \caption{\textbf{Quantitative ablation studies on several novel designs in our method.} We color code the \best{\textbf{best}}. The results indicate that our full model achieves the best performance.}
    \label{tab:quan_ablation}

\vspace{-2em}
\end{table}

\bibliographystyle{eg-alpha-doi} 
\bibliography{main}


\end{document}